\documentclass[review]{elsarticle}

\usepackage{lineno, hyperref, amsmath, multirow, epstopdf}

\usepackage{amsmath} 		
\usepackage{amssymb}		
\usepackage{subfigure}		
\usepackage{amsthm}
\usepackage{url}
\usepackage{mdframed}
\usepackage{graphics} 		
\usepackage{epsfig} 			
\usepackage{color}

\newtheorem{remark}{Remark}

\modulolinenumbers[5]

\journal{Robotics and Autonomous Systems}

\begin{document}

\begin{frontmatter}

\title{Contact Modelling and Tactile Data Processing for Robot Skin}

\author{Wojciech Wasko}
\author{Alessandro Albini}
\author{Perla Maiolino}
\author{Fulvio Mastrogiovanni\fnref{mark:corresponding}}
\fntext[mark:corresponding]{Corresponding author. \textit{Email address}: \texttt{fulvio.mastrogiovanni@unige.it}}
\author{Giorgio Cannata}

\address{
W. Wasko is with Nvidia Corporation, Gdansk, Poland.\\
A. Albini, F. Mastrogiovanni and G. Cannata are with the Department of Informatics, Bioengineering, Robotics and Systems Engineering, University of Genoa, Italy.\\
P. Maiolino is with the Oxford Robotics Institute, Department of Engineering Science, University of Oxford, UK.}

\begin{abstract}

Tactile sensing is a key enabling technology to develop complex behaviours for robots interacting with humans or the environment.
This paper discusses computational aspects playing a significant role when extracting information about contact events.
Considering a large-scale, capacitance-based robot skin technology we developed in the past few years, we analyse the classical Boussinesq-Cerruti's solution and the Love's approach for solving a distributed inverse contact problem, both from a qualitative and a computational perspective.
Our contribution is the characterisation of algorithms performance using a freely available dataset and data originating from surfaces provided with robot skin.

\end{abstract}


\end{frontmatter}

\linenumbers

\section{Introduction}
\label{sec:introduction}

The problem of characterising the physical interaction between robots and humans or the environment, typically using an artificial \textit{sense of touch}, has received increasing attention in
the literature \cite{Dahiyaetal2010}.
Extensive work has been done to allow robots to obtain information about contact events \cite{Schmitz2011}.
Large-scale, whole-body robot skin is a key enabling technology to implement interactive robot behaviours, specifically to provide control algorithms with reliable information about contact features \cite{DelPreteetal2011, Deneietal2015, Youssefietal2015a}.

Two classes of approaches for obtaining meaningful information about contact events can be identified.
The first adopts data-driven, machine learning frameworks to deal with situations where it is not trivial to model the underlying transduction principles.
As pointed out in \cite{Seminaraetal2015}, data-driven approaches are appealing when modelling contact events is complex and it is difficult to model the sensor's response, specifically to take noise into account. 
The second focuses on appropriately modelling physical laws. 
Model-driven approaches are adopted whenever a (possibly simplified) model of the force distribution is available. 
Such models are usually based on principles of contact mechanics, and are aimed at determining closed-form solutions for contact shape reconstruction \cite{Philips1981, Fearing1985, Howe1993}.
As a consequence, the two approaches originate qualitatively different results, i.e., typically a class  \textit{label} in the first case or a contact \textit{shape} in the second.

For both data-driven and model-based approaches, three requirements must be considered:
\begin{enumerate}
\item[$R_1$] \textit{Generalisation}: run-time contact reconstruction or labelling should not assume any \textit{a priori} model of the object in contact.
\item[$R_2$] \textit{Efficiency}: the reconstruction or classification algorithm's execution time should be predictable.
\item[$R_3$] \textit{Scalability}: the solution of the reconstruction or labelling problem should scale to different contact area sizes.
\end{enumerate}

\begin{table}[!t]
\begin{center}
\scriptsize
\caption{Classification of selected contact modelling approaches.}
\label{tab:comparison}
\begin{tabular}{l | c c | c c c | c}
Reference									& d-d		& m-b		& $R_1$		& $R_2$		& $R_3$		& $\mathcal{O}$	\\
\hline
Kim \textit{et al}. \cite{Kim2005}					& \checkmark	& 			&			& \checkmark	& \checkmark	& $c^2$			\\
Goger \textit{et al}. \cite{Goger2009} 				& 			& \checkmark	& \checkmark	& \checkmark	&			& $ndc$			\\
Tawil \textit{et al}. \cite{Tawil2011}				&			& \checkmark	& \checkmark	& \checkmark	&			& $d^2$			\\
Drimus \textit{et al}. \cite{Drimus2011}			& \checkmark	& 			& \checkmark	& 			&			& $ndc$			\\
Decherchi \textit{et al}. \cite{Decherchi2011}		& \checkmark	&			&			& \checkmark	&			& $nf$			\\
Liu \textit{et al}. \cite{Liuetal2012}				& 			& \checkmark	&			& \checkmark	& 			& $n$			\\
Bhattacharjee \textit{et al}. \cite{Bhattacharjee2012}	& \checkmark	& 			& 			&			& \checkmark	& $ndc$			\\
Ho \textit{et al}. \cite{Ho2012}					&			& \checkmark	& \checkmark	& 			&			& $d^2+d^2log(d)$	\\
Xu \textit{et al}. \cite{Xu2013}					& \checkmark	& 			& 			& \checkmark	& 			& $d^2$			\\
Muscari \textit{et al}. \cite{Muscari2013}			&			& \checkmark	& \checkmark	& \checkmark	& \checkmark 	& $d^2$			\\
Seminara \textit{et al}. \cite{Seminaraetal2015}		& 			& \checkmark	& \checkmark	& 			& \checkmark	& $d^2$			\\
\end{tabular}
\end{center}
\end{table}
Reconstructing the contact shape on the robot's surface or classifying it to inform robot behaviours require finding a trade-off between all requirements above, which may be in contrast with each other.
They have been considered only to a limited extent in the literature.
Table \ref{tab:comparison} reports selected data-driven and model-based approaches considering such an interplay.
It indicates the worst-case computational complexity of each method using Big $\mathcal{O}$ notation, where $c$ is the number of classes to discriminate from, $n$ the size of the training set, $d$ the data size, and $f$ is the number of features.
With the sole exception of \cite{Muscari2013}, all approaches target a subset of the requirements, and do not consider them as a whole. 
If we restrict the analysis to the approaches aimed at reconstructing the contact \textit{shape} and considering requirements $R_1$, $R_2$ or $R_3$, only the work in \cite{Muscari2013, Seminaraetal2015} appears relevant.  

It is necessary to better characterise the interplay between generalisation, computational efficiency and scalability when solving the problem of reconstructing the contact shape using tactile information.
The contribution of this paper is a discussion about the application of foundational model-based approaches, such as the Boussinesq-Cerruti \cite{SvecandGladwell1971, LiandBerger2001} and the Love solutions \cite{Love1927, Love1929, BeckerBevis2004} to the problem of reconstructing the contact \textit{shape} when large-scale, capacitance-based robot skin is used, from a generalisation, computational efficiency and scalability perspectives.
A software framework with algorithms to perform specific tests is available online as open source software\footnote{Webpage: http://git.io/contact-modelling.}.

The paper is organised as follows.
Section \ref{sec:prob} describes the problem we want to solve using the Boussinesq-Cerruti and the Love formulations.
Reconstruction algorithms are described in Section \ref{sec:reconstruction}.
A discussion about the physical plausibility of solutions is reported in Section \ref{sec:plausibility}.
Implementation details and performance issues are described in Section \ref{sec:exp_validation}.
Conclusions follow. 

\begin{figure}[t!]
\centering
\includegraphics[width =\columnwidth]{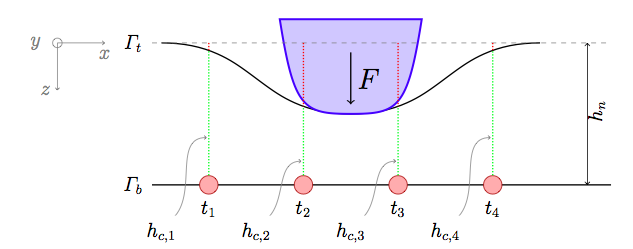}
\caption{An ideal cross-section of ROBOSKIN in a contact event: the top layer is deformed by a pressure distribution, which is measured by taxels $t_i.$}
\label{fig:contact}
\end{figure}

\section{Assumptions and Problem Statement}
\label{sec:prob}

We target the ROBOSKIN technology \cite{Schmitz2011, Billardetal2013}.
A single transducer (i.e., a \textit{taxel}) $t_i$ is a layered structure: the bottom layer $\Gamma_b$ is a positive electrode, the top layer $\Gamma_t$ is a ground electrode, and the mid layer is a soft elastomeric material \cite{Maiolinoetal2015}, see Figure \ref{fig:contact}.
In ROBOSKIN, taxels are spatially arranged in $3$ $cm$ side triangular modules, which can be connected to form large \textit{patches}.
Each module hosts $12$ taxels.
A normal force $F$ exerted on $\Gamma_t$ produces variations in each taxel capacitance:
\begin{equation}
\Delta C_i = C_{c,i} - C_n = \epsilon_{0}\epsilon_{r}A\frac{h_n - h_{c,i}}{h_{c,i} h_n} 
\label{eq:delta_cap}
\end{equation}
where $C_{c,i}$ and $h_{c,i}$ are, respectively, the capacitance value and the elastomer thickness for $t_i$ when \textit{contact} occurs, $C_n$ and $h_n$ correspond to the \textit{no contact} or nominal case, $\epsilon_0$ is the dielectric constant, $\epsilon_r$ is the relative static permittivity, and $A$ is the taxel's area.
Contact induces an increase in capacitance, i.e., $\Delta C_i > 0$ corresponds to a contact event. 

We are interested in solving an Inverse Elastic Problem (IEP), i.e., reconstructing the contact shape originating from the application of a force distribution \cite{Johnson1987}.
This implies determining the shape of $\Gamma_t$ (in terms of surface tractions) using taxel readings in \eqref{eq:delta_cap}.
We resort to the theory of linear elasticity, and we pose the following assumptions \cite{Slaughter2002}: (i) there exists a linear relation between stress and strain (deformation), and (ii) strains are infinitesimal.
As discussed in \cite{Maiolinoetal2015}, these assumptions are reasonable for ROBOSKIN: stress-strain relationships are shown to be piecewise linear for different elastomeric materials.
Therefore, the superposition principle can be applied, i.e., the strain resulting from a set of stresses is given by the sum of strains caused by individual stresses.
Solutions to problems grounded in the theory of linear elasticity deal with elastic half-spaces, i.e., solids bounded by a plane conventionally defined by $z=0$.
Due to its finite thickness, ROBOSKIN cannot be treated as an open half-space.
However, using the superposition principle, strains can be modelled using \textit{effective surface displacements}.
For a given taxel $t_i$, if $u(\sigma_i)$ is the deformation caused by the stress $\sigma_i$, the corresponding effective surface displacement $\delta_i$ is: 
\begin{equation}
\delta_i = u(\sigma_i)|_{z=0} - u(\sigma_i)|_{z=h_{c,i}}
\label{eq:esd}
\end{equation}
where $h_{c,i}$ is the elastomer thickness for taxel $t_i$. 
Therefore, if $Q \in R^M$ is a vector of surface tractions and $D \in R^N$ a vector of displacements, IEP consists in finding a function $g$ such that:
\begin{equation}
Q = g(D)
\end{equation}
In our case, the function $g$ is linear and requires the inversion of a matrix $C \in R^{M \times N}$. 
The use of displacements allows us to solve IEP using the Boussinesq-Cerruti or the Love formulations.
Both formulations are well-known, and practical information about to solve them, including how to invert $C$, have been discussed elsewhere \cite{Johnson1987, Seminaraetal2015}.
Here, we focus on the assumptions needed to implement algorithms considering the requirements introduced above. 

\textit{Normal forces}.
Since ROBOSKIN adopts capacitance-based sensors, it can detect only normal forces.
While this limits the type of contact events that can be detected, it allows for the simplification of IEP.
Only a subset of tractions (respectively, displacements) are to be included in $Q$ (respectively, in $D$), i.e., the elements of $C$ related to tractions along the $x$ and $y$ axes in the elastic half-space can be explicitly set to $0$.
As discussed in \cite{Seminaraetal2015}, this makes $C$ a sparse matrix, which can be efficiently inverted in $\mathcal{O}(M^2N)$ \cite{Lietal2013}. 

\textit{Taxels layout}.
The solution to IEP must be independent from the underlying robot skin taxels layout, which depends on design and manufacturing considerations \cite{Anghinolfietal2013}.
We discretise the IEP domain in an array of spatially distributed tractions and displacements by means of a virtual grid.
Each grid cell contains a discretised value corresponding to tractions $Q$ or displacements $D$ in the cell's location.
Although it is not explicitly required in our approach, we will consider grids in $\mathbb{R}^2$.
This requires us to obtain a $2D$ representation of the robot's surface provided with robot skin.
Methods to achieve this have been discussed in \cite{Cannataetal2010b, Deneietal2015}.
It is noteworthy that grid cells are the codomain of a chart whose domain is in $3D$ space, and in this case the manifold is the robot's surface.

\section{On the Reconstruction of Contact Shapes from Robot Skin Measurements}
\label{sec:reconstruction}

In this work, we are interested in solving an IEP in which surface tractions $Q$ represent the contact shape, and displacements $D$ are given as a function of taxel's measurements in the form of \eqref{eq:delta_cap}. 
First, we introduce a solution based on the Boussinesq-Cerruti model \cite{SvecandGladwell1971, LiandBerger2001}, then we point out its limitations, and finally we introduce the Love's model \cite{Love1927, Love1929, BeckerBevis2004}, which we adapt to our robot skin design.

\subsection{Boussinesq-Cerruti's Formulation}
\label{sec:Boussinesq_solution}
\textit{Derivation of Influence Coefficients}.
The formulation of IEP according to the Boussinesq-Cerruti solution assumes the displacements in the elastomeric layer to be caused by a distribution of concentrated forces acting at discrete locations on its surface \cite{Johnson1987}.
In particular, the Boussinesq-Cerruti solution for IEP is related to both concentrated normal and tangential forces $F$.
For elastomeric materials whose Poisson's ratio is equal to $\nu = 0.5$, which is a good approximation for the elastomer used in the employed ROBOSKIN prototypes, namely \textit{Ecoflex} \cite{Maiolinoetal2015}, the original formulas determining deformation contributions along $x$, $y$ and $z$ simplify significantly to:
\begin{align}
u^{x}& = \frac{3}{4\pi E} \left[ F^{x} \left(\frac{1}{\rho} + \frac{x^{2}}{\rho^{3}} \right) + F^{y} \frac{xy}{\rho^{3}} + F^{z}\frac{xz}{\rho^{3}}\right]
\label{eq:ux}
\\
u^{y}& = \frac{3}{4\pi E} \left[ F^{x} \frac{xy}{\rho^{3}} + F^{y} \left(\frac{1}{\rho} + \frac{y^{2}}{\rho^{3}} \right) + F^{z}\frac{yz}{\rho^{3}} \right]
\label{eq:uy}
\\
u^{z}& = \frac{3}{4\pi E} \left[ F^{x} \frac{xz}{\rho^{3}} + F^{y} \frac{yz}{\rho^{3}} + F^{z} \left( \frac{1}{\rho} + \frac{z^{2}}{\rho^{3}} \right) \right]
\label{eq:uz}
\end{align}
where:
\begin{equation}
\rho = \sqrt{x^{2}+y^{2}+z^{2}}
\label{eq:rho}
\end{equation}
and $E$ is the Young's modulus of the dielectric material.

Tractions vector $Q$ consists of $N = 3L$ components $q_j$ of $L$ concentrated forces $F_l$, and each concentrated force has three spatial components, respectively along $x$, $y$ and $z$:
\begin{equation}
Q = [F^{x}_{1} F^{y}_{1} F^{z}_{1} \ldots F^{x}_{l} F^{y}_{l} F^{z}_{l} \ldots F^{x}_{L} F^{y}_{L} F^{z}_{L}]^{T}
\label{eq:explodedQ}
\end{equation}
These concentrated forces result in a vector of displacements $D$ along $x$, $y$ and $z$ at $M = 3K$ distinct locations:
\begin{equation}
D = [\delta^{x}_{1} \delta^{y}_{1} \delta^{z}_{1} \ldots \delta^{x}_{k} \delta^{y}_{k} \delta^{z}_{k} \ldots \delta^{x}_{K} \delta^{y}_{K} \delta^{z}_{K}]^{T}
\label{eq:D}
\end{equation}
If we consider effective surface displacements \eqref{eq:esd}, a $3 \times 3$ sub-matrix of $C$ containing influence coefficients, which relate effective displacements at location $k$ with the forces applied at location $l$, is determined:

\begin{equation}
C_{\bar{k}, \bar{l}} =  
\begin{pmatrix}
c_{3k,3l} & c_{3k,3l+1} &  c_{3k,3l+2} \\
c_{3k+1,3l} & c_{3k+1,3l+1} & c_{3k+1,3l+2} \\
c_{3k+2,3l} & c_{3k+2,3l+1} & c_{3k+2,3l+2} 
\end{pmatrix}
\label{eq:coeff_new}
\end{equation}
where $\bar{k}$ ranges between $3k$ and $3k+2$, and $\bar{l}$ ranges between $3l$ and $3l+2$.
Influence coefficients in \eqref{eq:coeff_new} can be expressed in terms of geometric and physical properties of the elastomer, using \eqref{eq:uy} and \eqref{eq:uz}: 

\begin{equation*}
c_{3k,3l} = \frac{3}{4\pi E}\left[\frac{2x^{2}_{kl}+y^{2}_{kl}}{\left(\sqrt{x^{2}_{kl}+y^{2}_{kl}}\right)^{3}}-\frac{2x^{2}_{kl}+y^{2}_{kl}+h_c^{2}}{\left(\sqrt{x^{2}_{kl}+y^{2}_{kl}+h_c^{2}}\right)^{3}}\right]
\end{equation*}
\begin{equation*}
c_{3k,3l+1}=\frac{3}{4\pi E}\left[\frac{x_{kl}y_{kl}}{\left(\sqrt{x^{2}_{kl}+y^{2}_{kl}}\right)^{3}}-\frac{x_{kl}y_{kl}}{\left(\sqrt{x^{2}_{kl}+y^{2}_{kl}+h_c^{2}}\right)^{3}}\right]
\end{equation*}
\begin{equation*}
c_{3k,3l+2}=\frac{3}{4\pi E}\left[-\frac{x_{kl}h_c}{\left(\sqrt{x^{2}_{kl}+y^{2}_{kl}+h_c^{2}}\right)^{3}}\right]
\end{equation*}
\begin{equation*}
c_{3k+1,3l}=\frac{3}{4\pi E}\left[\frac{x_{kl}y_{kl}}{\left(\sqrt{x^{2}_{kl}+y^{2}_{kl}}\right)^{3}}-\frac{x_{kl}y_{kl}}{\left(\sqrt{x^{2}_{kl}+y^{2}_{kl}+h_c^{2}}\right)^{3}}\right]
\end{equation*}
\begin{equation*}
c_{3k+1,3l+1}=\frac{3}{4\pi E}\left[\frac{x^{2}_{kl}+2y^{2}_{kl}}{\left(\sqrt{x^{2}_{kl}+y^{2}_{kl}}\right)^{3}}-\frac{x^{2}_{kl}+2y^{2}_{kl}+h_c^{2}}{\left(\sqrt{x^{2}_{kl}+y^{2}_{kl}+h_c^{2}}\right)^{3}}\right]
\end{equation*}
\begin{equation*}
c_{3k+1,3l+2}=\frac{3}{4\pi E}\left[-\frac{y_{kl}h_c}{\left(\sqrt{x^{2}_{kl}+y^{2}_{kl}+h_c^{2}}\right)^{3}}\right]
\end{equation*}
\begin{equation*}
c_{3k+1,3l}=\frac{3}{4\pi E}\left[-\frac{x_{kl}h_c}{\left(\sqrt{x^{2}_{kl}+y^{2}_{kl}+h_c^{2}}\right)^{3}}\right]
\end{equation*}
\begin{equation*}
c_{3k+2,3l+1}=\frac{3}{4\pi E}\left[-\frac{y_{kl}h_c}{\left(\sqrt{x^{2}_{kl}+y^{2}_{kl}+h_c^{2}}\right)^{3}}\right]
\end{equation*}
\begin{equation*}
c_{3k+2,3l+2}=\frac{3}{4\pi E}\left[\frac{1}{\sqrt{x^{2}_{kl}+y^{2}_{kl}}}-\frac{x^{2}_{kl}+y^{2}_{kl}+2h_c^{2}}{\left(\sqrt{x^{2}_{kl}+y^{2}_{kl}+h_c^{2}}\right)^{3}}\right]
\end{equation*}
where the term $x_{kl}$ (respectively, $y_{kl}$) is the distance between the application point of force $F_l$ and the location of the displacement $\delta_k$ projected onto the $x$ (respectively, $y$) axis.

\textit{Limitations of the Boussinesq-Cerruti's Solution}.
A brief analysis of the coefficients in \eqref{eq:coeff_new} shows that there exists a singularity in the solution for $x_{kl} = y_{kl} = 0$.
Specifically, five coefficients become infinite for that choice of parameters.
Geometrically, this choice corresponds to the situation in which the point of application of a force vector $F_l$ on the robot skin and the measurement's location of the displacement $\delta_k$ coincide.

\begin{figure}[t]
\centering
\includegraphics[width =\columnwidth]{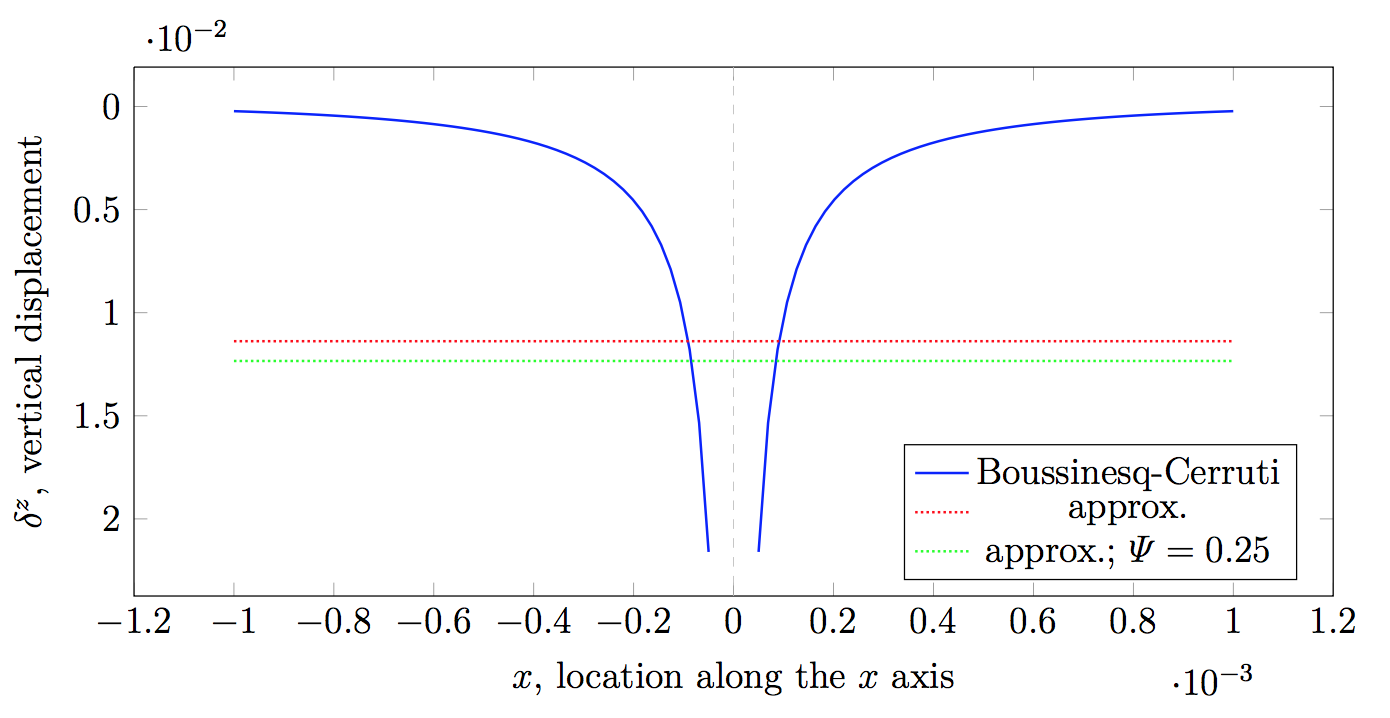}
\caption{Singularity of the Boussinesq-Cerruti solution. The solution is presented along the $x$ axis with $y = z = 0$, for a force $F^{z} = 1$ $N$, elastic modulus $E = 2.1e5$ $Pa$, robot skin thickness $h_{n} = 2$ $mm$, and size of the grid cell $dx = dy = 0.2$ $mm$. It is noteworthy that the theoretical displacement at $x = y = z = 0$ is infinite.
}
\label{fig:approx}
\end{figure}

To overcome this singularity issue, an approximate solution has been proposed in \cite{Muscari2013}\footnote{The interested reader is referred to the referenced publication for more details. Only a conceptual overview is given here.}. 
In order to obtain it, Muscari and colleagues pretend that the employed robot skin has a thickness of $h_c + z_{0,h_c}$ where $z_{0,h_c}$ is a small offset.
Forces are considered to be applied to the top of the \emph{approximated} robot skin, and the corresponding strain is computed using the work-force theorem and the Hooke's law.
The final form for the generic displacement $\delta_k$ of the approximate solution is:
\begin{align}
\delta_k^{x}&=\frac{9}{4\pi E}\frac{F_l^{x}}{z_{0,h_c}}\Psi\left(\frac{h_c}{z_{0,h_c}}\right)
\\
\delta_k^{y}&=\frac{9}{4\pi E}\frac{F_l^{y}}{z_{0,h_c}}\Psi\left(\frac{h_c}{z_{0,h_c}}\right)
\\
\delta_k^{z}&=\frac{9}{2\pi E}\frac{F_l^{z}}{z_{0,h_c}}\Psi\left(\frac{h_c}{z_{0,h_c}}\right)
\label{eq:delta_approx}
\end{align}
where $\Psi(x) = x^{-1} (0.2431 x - 0.1814)$ and:
\begin{equation}
z_{0,h_c} = \sqrt{\frac{3}{2\pi} A(s_{k})}
\end{equation}
with $A(s_{k})$ is the area over which the concentrated force $F_l$ is exerted, i.e., in our implementation taken to be equal to the area of a grid cell $s_k$.
In the original implementation, it is taken that since $z_{0, h_c} \ll h_c$, the $\Psi(x) \approx 0.25$ and it is explicitly set to this value (see Figure \ref{fig:approx} for a comparison).

\begin{figure}[t]
\centering
\includegraphics[width =\columnwidth]{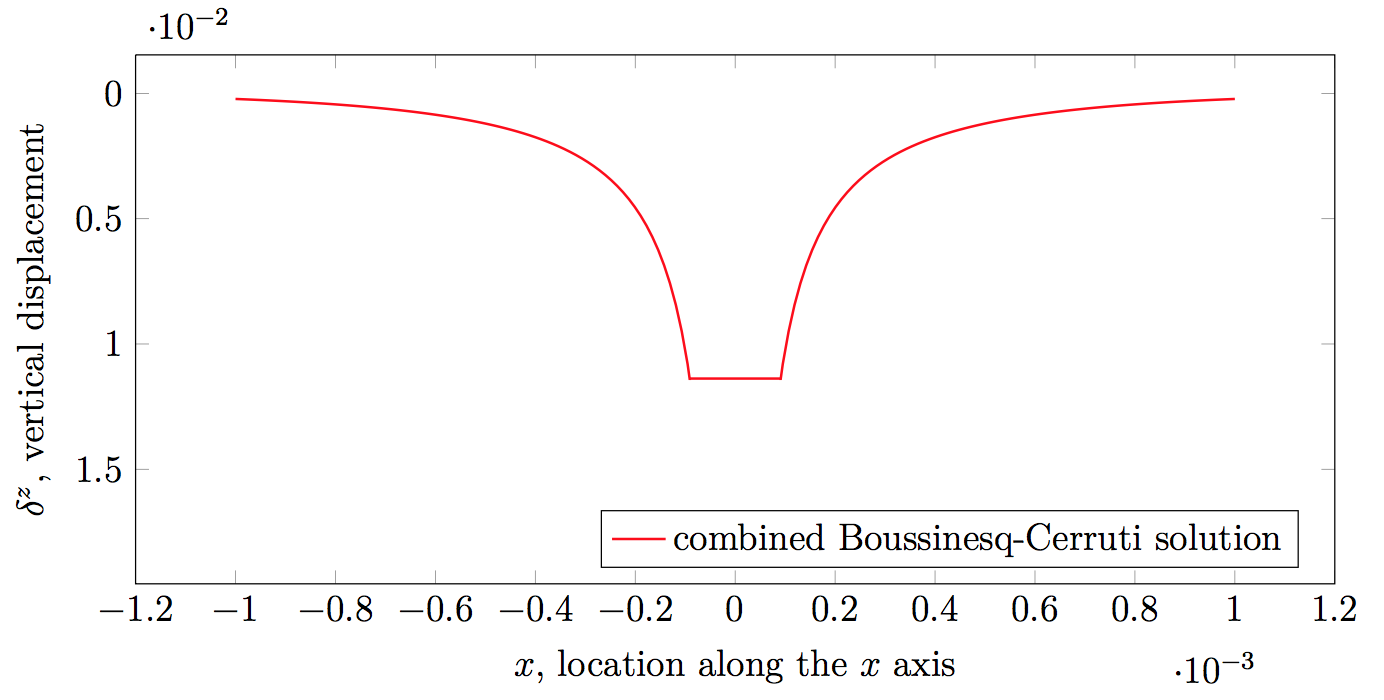}
\caption{Continuous solution for the Boussinesq-Cerruti singularity issue.}
\label{fig:combined}
\end{figure}

It is noteworthy that Muscari and colleagues consider only regular grids with equal grid cells, and therefore assume a uniform spacing between the nodes representing points of application of force and sensing locations.
Since these nodes are placed perfectly one over another, they apply the approximate solution when computing influence coefficients for nodes that correspond to the same point in space.
In other cases, it is obvious that either one of $x_{kl}$ or $y_{kl}$ is non-zero, and therefore the solutions are not singular.
It logically follows that the criterion the authors apply for choosing between the direct Boussinesq-Cerruti solutions and the approximated solution is discrete in nature.
In order to apply the Boussinesq-Cerruti solutions to any combination of (not necessarily regular) grids, with arbitrary cell sizes, a \emph{continuous} criterion must be employed instead.
To this end, one must consider the nature of the functions constituting the Boussinesq-Cerruti solutions.
The singularity occurs with $z = 0$, and therefore the problem is planar in nature, i.e., solutions are functions of $(x_{kl}, y_{kl})$.
If these functions could be shown to be concave (i.e., their Hessian matrices were positive semi-definite), then it would be trivial to choose a solution (Figure \ref{fig:combined}).
Unfortunately, this cannot be shown trivially.
If we consider the expression of the influence coefficient $c_{3k+2, 3l+2}$ in \eqref{eq:coeff_new} for a generic displacement $\delta^{z}_j$ and force $F^{z}_k$, we express it adopting a polar coordinate system $\langle r^{2}, \phi \rangle$, where $r^2 = x^{2}_{kl} + y^{2}_{kl}$ and $\phi = atan2(y_{kl},x_{kl})$, and we try to find critical points by calculating:
\begin{equation*}
\frac{\partial c_{3k+2,3l+2}}{\partial r} = 0
\end{equation*}
it can be shown by applying the Sturm's theorem to the resulting polynomial that there exists a critical point in the interval $r \in (0, \infty)$.
However, determining the nature of the critical point would require us to solve a quartic polynomial and possibly further steps to examine growing orders of derivatives.
This approach was deemed unsatisfactory, but no better continuous criterion can be proposed at this time.
For this reason, the aforementioned criterion is what the current implementation employs, i.e., both the original Boussinesq-Cerruti and the approximated solutions are computed, and then the one with the lower absolute value is used for the influence coefficient.

\subsection{Love's Formulation}
\label{sec:Love_solution}

\textit{Derivation of Influence Coefficients}.
The formulation of the elastic problem for the Love's solution considers displacements $\delta_k$ in the elastomeric material to be determined by a distribution of \emph{normal} pressures acting over rectangular areas on the robot's surface \cite{Love1929}. 
Such areas, or grid cells, have sizes $2a$ and $2b$ (Figure \ref{fig:contact}).

\begin{figure}[t]
\centering
\includegraphics[width = \columnwidth]{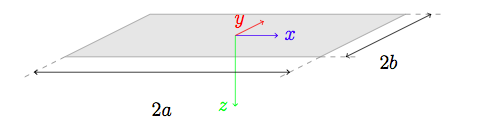}
\caption{Love'€™s solution assumes rectangular pressure cells at the surface of an elastic half-space.}
\label{eq:contact}
\end{figure}
For the sake of notation, let us assume that a point on the boundary of the half-space (i.e., the robot skin's surface $\Gamma_t$) has Cartesian coordinates $\langle x', y', 0 \rangle$ and a point within the elastomer has Cartesian coordinates $\langle x, y, z \rangle$ with $z > 0$. Furthermore, let $r$ be an auxiliary variable such that $r^{2} = \Delta x^{2} + \Delta y^{2} + z^{2}$, where $\Delta x = x' - x$ and $\Delta y = y' - y$.
Therefore, solutions follow the notion of spatial derivatives of elastic potential functions introduced by Boussinesq, whereas deformations are computed as:
\begin{align}
u^{x}& = -\frac{1}{4\pi} \left[ \frac{2(1+\nu)(1-2\nu)}{E}\frac{\partial \chi}{\partial x}+\frac{2(1+\nu)z}{E}\frac{\partial V }{\partial x} \right]
\\
u^{y}& = -\frac{1}{4\pi} \left[ \frac{2(1+\nu)(1-2\nu)}{E}\frac{\partial \chi}{\partial y}+\frac{2(1+\nu)z}{E}\frac{\partial V }{\partial y} \right]
\\
u^{z}& = -\frac{1}{4\pi} \left[ \frac{4(1-\nu^{2})(1-2\nu)}{E} V + \frac{2(1+\nu)z}{E}\frac{\partial V }{\partial z} \right]
\end{align}
where $\chi$ and $V$ are two functions describing the elastic potentials, defined respectively as:
\begin{equation}
\chi = \int\limits_{-a}^a \int\limits_{-b}^b p\log(z+r)dx'dy'
\label{eq:chi}
\end{equation}
and:
\begin{equation}
V = \int\limits_{-a}^a \int\limits_{-b}^b pr^{-1}dx'dy'
\label{eq:V}
\end{equation}
where $p$ is the pressure acting on the grid cell.
The term $\chi$ in \eqref{eq:chi} is called Boussinesq's $3D$ elastic logarithmic potential, while the term $V$ in \eqref{eq:V} is the Newtonian potential of the surface distribution.
As described in \cite{BeckerBevis2004}, integrals in \eqref{eq:chi} and \eqref{eq:V} can be evaluated to obtain closed-form solutions for the deformations in an elastic half-space excited by a \emph{uniform normal pressure} applied over a rectangular grid cell, as follows:
\begin{align}
u^{x}& = -\frac{p}{4\pi}\left[\frac{(1+\nu)(1-2\nu)}{E}(J_{2}-J_{1}) + \right.\nonumber
\\
&\left.\qquad\qquad\qquad \frac{2(1+\nu)z}{E}\ln\left(\frac{\Delta y +r_{20}}{\Delta y + r_{10}}\right)\right]^{y'=b}_{y'=-b}
\\
u^{y}& = -\frac{p}{4\pi}\left[\frac{(1+\nu)(1-2\nu)}{E}(K_{2}-K_{1}) + 
\right.\nonumber
\\
&\left.\qquad\qquad\qquad \frac{2(1+\nu)z}{E}\ln\left(\frac{\Delta x +r_{02}}{\Delta x + r_{01}}\right)\right]^{x'=a}_{x'=-a}
\\
u^{z}& = \frac{p}{4\pi}\left\{\frac{4(1-\nu^{2})}{E}(L_{1}-L_{2}) +
\right.\nonumber
\\
&\left. z\left[\tan^{-1}\left(\frac{(a-x)\Delta y}{zr_{10}}\right)+\tan^{-1}\left(\frac{(a+x)\Delta y}{zr_{20}}\right)\right]\right\}^{y'=b}_{y'=-b}
\end{align}
where:
\begin{align}
J_{j}& = \Delta y\left[\ln(z+r_{j0})-1\right] + z \ln\left(\frac{1+ \psi_{j0}}{1- \psi_{j0}}\right) + \nonumber
\label{eq:Jj}
\\
&\qquad\qquad\qquad\qquad 2|a \mp x| \tan^{-1}\left(\frac{|a \mp x|\psi_{j0}}{z+\beta_{j0}}\right)
\\
K_{j}& = \Delta x\left[\ln(z+r_{0j})-1\right] + z \ln\left(\frac{1+ \psi_{0j}}{1- \psi_{0j}}\right) + \nonumber
\label{eq:Kj}
\\
&\qquad\qquad\qquad\qquad 2|b \mp y| \tan^{-1}\left(\frac{|b \mp y| \psi_{0j}}{z+ \beta_{0j}}\right)
\\
L_{j}& = \Delta y\left[\ln(\pm a - x +r_{j0})-1\right] + \nonumber
\label{eq:Lj}
\\
&(\pm a - x) \ln\left(\frac{1+ \psi_{j0}}{1- \psi_{j0}}\right) + 2z \tan^{-1}\left(\frac{z \psi_{j0}}{(\pm a - x) \beta_{j0}}\right)
\\
\psi_{j0}& = \frac{\Delta y}{r_{j0}+\beta_{j0}}
\\
\psi_{0j}& = \frac{\Delta x}{r_{0j}+\beta_{0j}}
\\
r_{j0}& = \sqrt{(a \mp x)^{2}+ \Delta y^{2} +z^{2}}
\\
r_{0j}& = \sqrt{\Delta x^{2} +(b \mp y)^{2}+ z^{2}}
\\
\beta_{j0}& = \sqrt{(a \mp x)^{2}+z^{2}}, \beta_{0j}=\sqrt{(b \mp y)^{2}+z^{2}}
\label{eq:Betaj0}
\end{align}
In \eqref{eq:Jj}-\eqref{eq:Betaj0}, the parameter $j$ determines which sign is used, i.e., the upper (respectively, lower) sign corresponds to $j = 1$ (respectively, $j = 2$), and $p$ is the value of the (uniform) pressure acting on the cell.

It is noteworthy that deformations are computed as linear functions of the pressures exerted on the robot skin and therefore the model is still linear, which allows us to take advantage of the superposition principle.
In the Love's formulation, the tractions vector $Q$ consists of $N$ normal components of a pressure distribution: 
\begin{equation}
Q = [p_{1} \ldots p_{n} \ldots p_{N}]^{T}
\end{equation}

These pressures result in a distribution of displacements $D$ along the $x$, $y$ and $z$ directions at $K$ locations in the same form of \eqref{eq:D}.
Therefore, the $3 \times 1$ sub-matrix of $C$ contains influence coefficients which relate displacements at location $k$ with pressures applied at location $n$. According to \eqref{eq:esd}, these can be expressed as:
\begin{equation}
C_{3k:3k+2,n} = \begin{bmatrix}
c_{3k,n}\\
c_{3k+1,n}\\
c_{3k+2,n}
\end{bmatrix}
\end{equation}
where:
\begin{align}
c_{3k,n}& = -\frac{p}{4\pi} \left\{ \frac{(1+\nu)(1-2\nu)}{E} \left[\left.(J_{2}-J_{1})\right|_{z=0}- \right.\right.
\nonumber
\\
&\left.\left. (J_{2}-J_{1})\right|_{z=h_c} \right] -\frac{2(1+\nu)h_c}{E}
\nonumber
\\
&\left. \ln \left( \left. \frac{\Delta y +r_{20}}{\Delta y +r_{10}}\right|_{z=h_c} \right) \right\}^{y'=b}_{y'=-b}
\\
c_{3k+1,n}& = -\frac{p}{4\pi} \left\{ \frac{(1+\nu)(1-2\nu)}{E} \left[ \left. (K_{2} - K_{1})\right|_{z=0} - \right. \right.
\nonumber
\\
&\left.\left. (K_{2} - K_{1})\right|_{z=h_c} \right] -\frac{2(1+\nu)h_c}{E}
\nonumber
\\
&\left. \ln \left( \left. \frac{\Delta y + r_{02}}{\Delta y + r_{01}}\right|_{z=h_c} \right) \right\}^{y'=b}_{y'=-b}
\\
c_{3k+2,n}& = -\frac{p}{4\pi} \left\{ \frac{4(1+\nu^{2})}{E} \left[ \left. (L_{1} - L_{2})\right|_{z=0} - \right.\right.
\nonumber
\\
&\left.\left. (L_{1}-L_{2})\right|_{z=h_c} \right] -
h_c \left( \tan^{-1}\frac{(a-x)\Delta y}{h_c r_{10}} + \right.
\nonumber
\\
&\left.\left. \tan^{-1}\frac{(a+x)\Delta y}{h_c r_{20}}\right)\right\}^{y'=b}_{y'=-b}
\end{align}
and terms $J_1$, $J_2$, $K_1$, $K_2$, $L_1$, $L_2$ are characterized by \eqref{eq:Jj}, \eqref{eq:Kj} and \eqref{eq:Lj}.

\begin{figure}[t]
\centering
\subfigure{
\includegraphics[width=0.9\columnwidth]{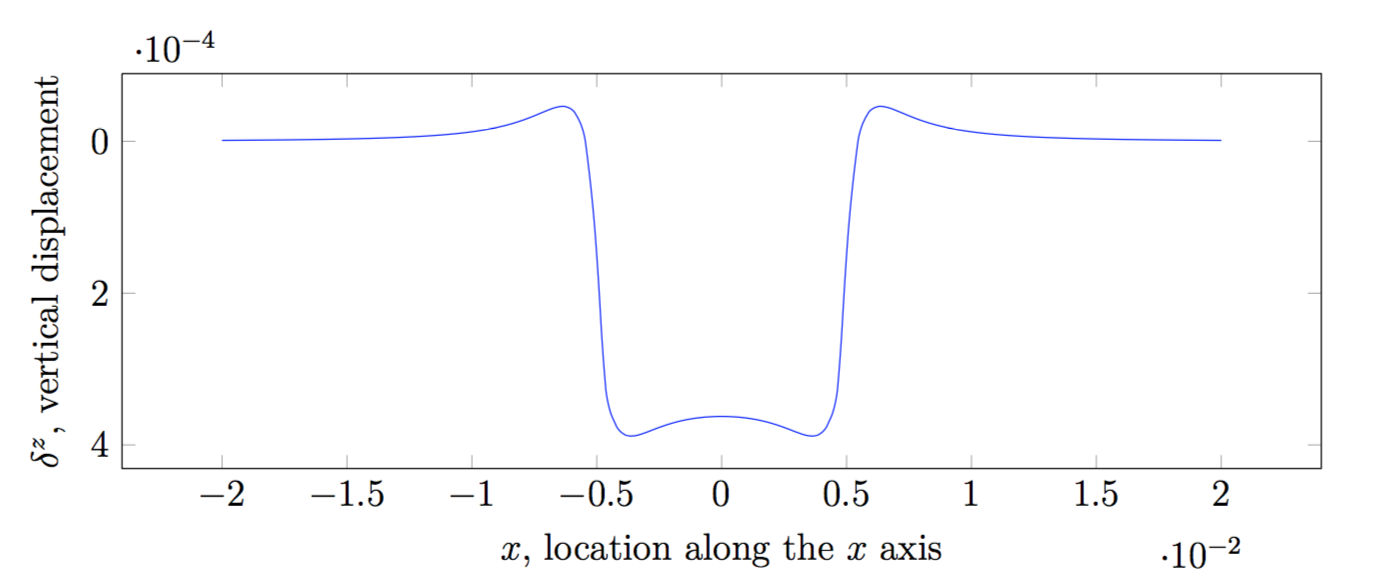}
\label{fig:LoveExample1}}
\subfigure{
\includegraphics[width=0.9\columnwidth]{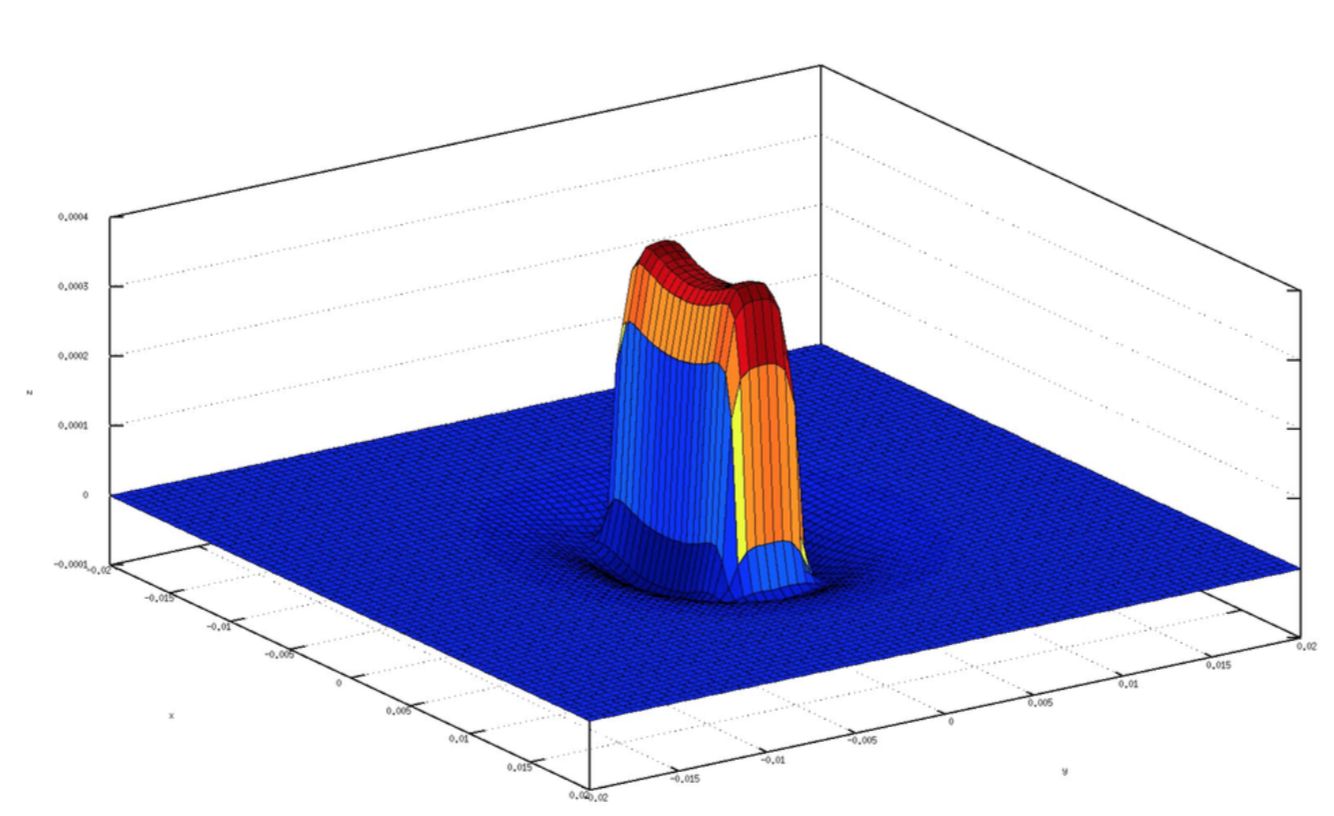}
\label{fig:LoveExample2}}
\caption{Love's solution for a uniform pressure $p = 1e5$ $Pa$ over a cell of $10 \times 4$ $mm$ size, with $E = 2.1e5$ $Pa$, $h_{n} = 2$ $mm$, shown along $y = 0$ (top), and the corresponding $3D$ view of the Love's solution (bottom).}
\label{fig:LoveExampleALL}
\end{figure}
Figure \ref{fig:LoveExampleALL} shows plots of solutions obtained for a uniform pressure exerted over a small grid cell.
On the top, the Figure shows a cross-section along $y = 0$, while on the bottom it contains a view of the normal deflection $\delta^{z}$.
If we compare the results in Figure \ref{fig:approx} and Figure \ref{fig:LoveExampleALL} on the top, it can be noticed that the Love's formulation is capable of modelling complex contact situations, including boundary effects, whereas the Boussinesq-Cerruti solution appears to simply smooth the contact shape, and is characterised by singularities.

\begin{figure}[t]
\centering
\includegraphics[width=0.9\columnwidth]{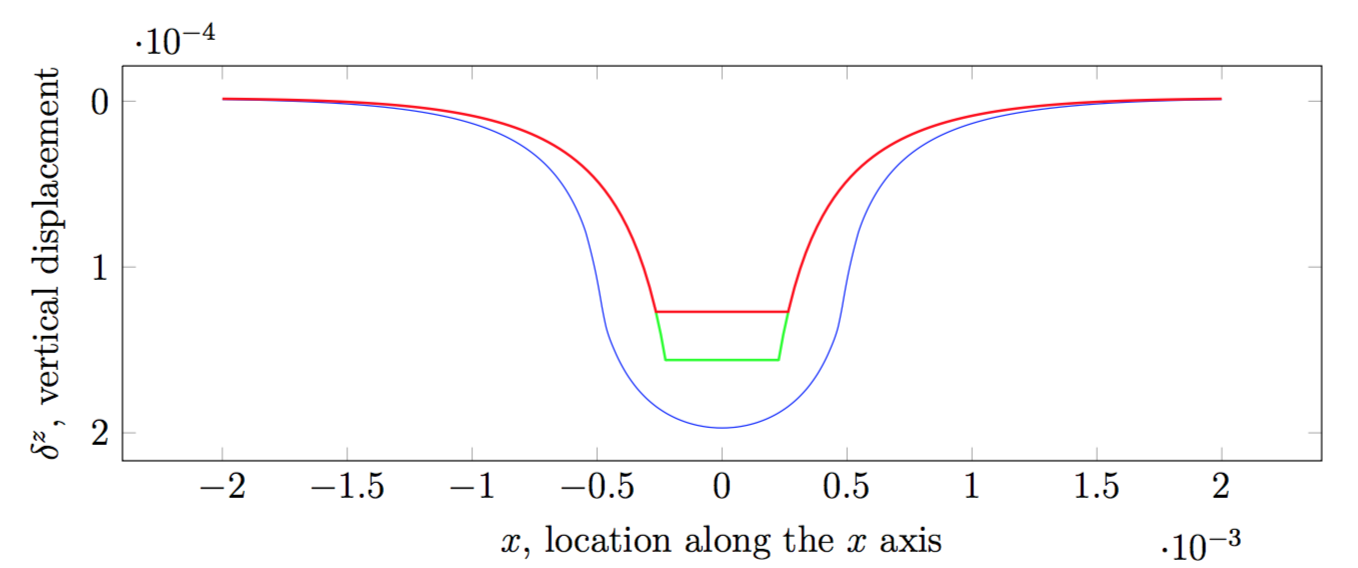}
\caption{Comparison between Boussinesq-Cerruti's solution (with exact computation of $\chi(x)$ in red and $\Psi(x) = 0.25$ in green, and Love's solution in blue for the same contact situation, with a pressure load $p = 1e5$ $Pa$ over a cell of $2a = 1  \times 2b = 0.4$ $mm$, force load $F^{z} =p\cdot2a\cdot2b = 0.04$ $N$, shown along $y=0$.}
\label{fig:comparison}
\end{figure}

\textit{Limitations of the Love's Solution}.
As it can be noticed in Figure \ref{fig:comparison}, if the size of the cell is decreased, the Love's solution \emph{smooths out} as well and becomes similar to the Boussinesq-Cerruti's solution, without the singularity.
This suggests that Love's formulation assures a certain degree of complexity required to model highly complex contact situations, specifically when the size of the grid cells is reduced.
It should also be noted that equations in the Love's formulation do not contain \emph{real} singularities, which is a fundamental feature to adopt the model with real-world robot skins.
As a matter of fact, there are expressions becoming singular in specific cases, but it can be shown that in all of them, they are multiplied by a coefficient which tends to $0$ faster than the expression becomes singular.
However, special care must be taken when implementing those equations, especially when dealing with finite precision arithmetic.


\subsection{Comparison between Boussinesq's and Love's Solutions}
\label{sec:comparison}

A qualitative comparison between the two solutions presented in Section \ref{sec:Boussinesq_solution} and Section \ref{sec:Love_solution} can be done by comparing deformations resulting from similar load conditions.
In order to generate \emph{similar} load conditions, we shall only consider loads normal to the surface. 
It is noteworthy that this is precisely the case with the robot skin technology we use, and due also to the fact that the Love's problem deals with normal pressures only.
Since the loads considered by the Boussinesq-Cerruti's and Love's formulations are different in nature (i.e., forces and pressures, respectively), we shall approximate the force acting in the Boussinesq-Cerruti model as:
\begin{equation}
F^{z}= p \cdot 2a \cdot 2b
\nonumber
\end{equation}
where $p$ is the normal pressure acting in the Love's formulation, and $2a$ and $2b$ are the sizes of the grid cell (where the load is applied).
Furthermore, for the approximate solution to the Boussinesq-Cerruti's model, we assume that the discretization of the forces space is done with elements such that the area over which the force is exerted is equal to that of the pressure's grid cell in the Love's model:
\begin{equation}
A(s) = 2a \cdot 2b
\nonumber
\end{equation}

A sample result of the comparison can be seen in Figure \ref{fig:comparison}.
The two models produce results within the same magnitude order, i.e., less than $2 \cdot 10^{-4}$, but nevertheless there is a significant difference in the amplitude of the deflection, the largest difference occurring at $x = y = 0$.
We can try to qualitatively compare this difference by computing the approximate Boussinesq-Cerruti's solution for a fixed force value $F^{z}$ and varying the area of the grid cell $A(s)$, and plotting it with a Love's solution for grid of size $2a = 2b = \sqrt{A(s)}$, and pressure value $p = F^{z} /(2a \cdot 2b)$, evaluated at $x = y = 0$.
The discrepancy in the two solutions is contrary to what one would expect.
Intuitively, the Boussinesq-Cerruti's solution (i.e., concentrated force) should result in an indentation with narrower, sharper slopes than the solution corresponding to the Love's problem.
On the other hand, the indentation in the point of application of the force would be expected to be \emph{deeper} in the Boussinesq-Cerruti's solution, since it originates from a \textit{concentrated} force, but this is not the case.

\section{Plausibility of Solutions}
\label{sec:plausibility}

By analysing the contact situation, one can derive a set of constraints on the model involved in the inverse elastic problem, which define the set of feasible solutions for tractions $Q$.
Usually, these constraints originate from the likelihood of observing some measurements due to the laws of physics governing the contact event.
It may occur that the feasible solution set does not overlap with the theoretical solutions obtained through the inversion of $C$.
This means that there might not exist a physically feasible solution (in terms of model parameters) which would produce \emph{exactly} the same data if a solution to the forward elastic problem was computed with the model's parameters taken to be such exact solutions.
This phenomenon is readily explained by taking into consideration the presence of noise in the data, which is the input to the inverse elastic problem.
Input data (e.g., taxel measurements) are affected by various sources of noise, most of which are unrelated to the nature of the elastomer employed in the robot skin, e.g., electric charge fluctuations on the capacitance-based taxels, limited processing precision of CDC units, or even the limited precision of computer arithmetic, just to name a few.
An exact solution to the inverse problem (if it exists) corresponds to modeling all these imperfections, which is usually not desirable.
One can therefore sacrifice the least-squares guarantee of a pseudo-inverse based solution for meeting physicality constraints on the tractions set.

Following the ideas presented in \cite{Seminaraetal2015}, one can assume that under common contact conditions, normal tractions (forces or pressures) can act only in a compressive way, i.e., $q_{j}^{z} > 0$ for every $j$.
Moreover, we assume that tangential tractions are generated exclusively by friction \cite{Murray1994}.
If we further constrain ourselves to single-contact cases only, then tangential tractions can be shown to be proportional to normal tractions.

\subsection{Compressive Normal Tractions with the Fourier-Motzkin Elimination}
\label{sec:compressive_FME}
There are many ways to constrain normal tractions to be compressive.
Let us assume that there are only normal tractions acting on the robot's body.
As a consequence, the relationship between displacements and tractions becomes:
\begin{equation}
D = C_{n} Q^{z}
\end{equation}
with a general solution in the form:
\begin{equation}
Q^{z} = C_{n}^{+}D + (I-C_{n}^{+}C_{n})z
\label{eq:genform}
\end{equation}
where $z \in \mathbb{R}^N$ such that $z = [z_1, \ldots, z_N ]^T$ is an arbitrary vector.
As mentioned above, constraining tractions to be non-negative can be expressed as the following system of inequalities:
\begin{equation}
\forall j \quad q_{j} > 0 
\end{equation}
If we restrict the domain in which to search for the non-negative solutions to the set of \emph{exact} solutions given by \eqref{eq:genform}, the problem of obtaining the non-negative constraints on the elements of $Q$ is equivalent to that of finding the components of the $z$ vector satisfying a system of linear inequalities:
\begin{equation}
\left(I - C_{n}^{+} C_{n}\right)
\begin{bmatrix}
z_1
\\
\vdots
\\
z_N
\end{bmatrix} \geq -C_{n}^{+}D
\nonumber
\end{equation}
It can be proved that a solution in closed form to this set of linear inequalities exists, and it can be obtained using a mathematical tool introduced by Fourier in \cite{Fourier1826}, later rediscovered by Lloyd L. Dines and Theodore S. Motzkin, and came to be known as the \textit{Fourier-Motzkin elimination}.

The process to obtain a solution is similar to that of Gauss€™ elimination, and consists in transforming a set of linear inequalities to an equivalent set, which is expressed without a subset of the original variables.
Without loss of generality and for the sake of conciseness, let us consider an arbitrary system of linear inequalities:
\begin{align}
a_{11}x_{1} + \cdots +&a_{1n}x_{n} \geq b_1
\nonumber
\\
\vdots
\nonumber
\\
a_{j1}x_{1} + \cdots +&a_{jn}x_{n} \geq b_j
\nonumber
\\
\vdots
\nonumber
\\
a_{m1}x_{1} + \cdots +&a_{mn}x_{n} \geq b_m
\nonumber
\end{align}
where the values of real coefficients $a_{ji}$ are not constrained, i.e., they can be positive, negative or equal to zero.
Then, we can eliminate a variable $x_e$ by transforming the original inequalities so that on the left-hand side they contain only $x_e$, and grouping them into three classes, depending on their direction (which corresponds to the sign of coefficients $a_{je}$):
\begin{enumerate}
\item $x_e \geq A_k, A_k = b'_{j}- \sum_{\substack{i\neq e}} a'_{ji}x_i$;
\item $x_e \leq B_l, B_l = b'_{j}- \sum_{\substack{i\neq e}} a'_{ji}x_i$;
\item inequalities where $x_e$ plays no role, denoted $\phi$.
\end{enumerate}
The original system of inequalities is then equivalent to:
\begin{equation}
\max(A_k) \leq \min(B_k) \wedge \phi
\nonumber
\end{equation}
where $b'_{j} = {b_j}/{a_{je}}$ and $a'_{ji} = a_{ji}/{a_{je}}$.

In the worst case (with an equal number of inequalities in the first and second group), the number of inequalities in a system created by eliminating one variable is ${n^2}/{4}$.
Running $p$ successive elimination steps will result in at most $4({n}/{4})^{2p}$ inequalities.
Many of these inequalities are redundant and the required number of inequalities can be shown to grow as a single exponential \cite{Monniaux2010}.

Considering again the inverse elastic problem, without enough insight, the Fourier-Motzkin elimination might seem promising, since it could be performed as an offline step, with back-substitution being the only online part.

\begin{remark}
The complexity of the Fourier-Motzkin algorithm must be analyzed.
Since the displacement vector $D$ is only known at run-time (as a consequence of a contact event detected by the robot skin), no detection of superfluous inequalities (as mentioned above) can be done offline.
Therefore, the worst-case size of the final system of inequalities obtained after eliminating all variables is:
\begin{equation}
N' = 4 \left(\frac{N}{4}\right)^{2N}
\end{equation}
which, for a moderately sized problem of reconstructing a distribution of $100$ tractions (e.g., a $10 \times 10$ grid) corresponds to approximately $1.6 \cdot 10^{280}$ inequalities.
Without elaborating the details, encoding this in an actual robot software architecture would require approximately $5.8 \cdot 10^{273}$ gigabytes of memory under double precision.

From this comparison, it should be clear that the Fourier-Motzkin elimination cannot be adopted in practice.
\end{remark}

\subsection{Compressive Normal Tractions with Non-negative Least Squares}
\label{sec:compressive_NNLS}
The problem of finding non-negative solutions to systems of linear equations is a well studied problem in mathematical optimization, driven mostly by engineering applications \cite{Chen2009}.
As it turns out, quite often physically feasible solutions are constrained to the non-negative only domain.

The first algorithm to find a least-squares solution in the non-negative domain is due to Lawson and Hanson \cite{Lawson1974}.
This algorithm belongs to the family of active set algorithms.
It tries to iteratively optimise a subset of system variables whose values can change (i.e., the free set) so that the final solution satisfies the non-negativity constraint with a minimal least square-error.
However, the performance of the algorithm proposed by Lawson and Hanson is hindered by the expensive computation of a matrix inversion related to the free subproblem at each refinement step of the solution.
Moreover, this method belongs to the \textit{single pivoting} group, meaning that at each iteration, only one variable is chosen to be moved between the free and the active sets, which increases the number of iterations required to reach a \emph{good} solution.
Since then, many other algorithms to compute the non-negative least-squares solution to a system of linear equations have been proposed, usually offering significant performance improvement over the Lawson-Hanson algorithm.
Most of the algorithms have been generalised to realise box-type constraints, such that each variable $x_i$ is subject to $l_i \leq x_i \leq u_i$ constraints.
A partial survey of these algorithms is given in \cite{Chen2009}.
The majority of the modern solvers belong to the group of \emph{block-pivoting}, \emph{active set} algorithms.
The main difference with respect to the original algorithm by Lawson and Hanson is that, in each iteration, the algorithms try to move more than one variable between the free and the active sets.
In general, the inverse elastic problem is \textit{dense} in its nature, meaning that (almost)
all elements of the influence coefficients matrix are non-zero.
Nevertheless, Mikael Adlers describes an efficient and robust algorithm for solving box-constrained linear problems which remain highly efficient even for dense problems \cite{Adlers1998}. This algorithm, known as BLOCK3 block pivoting algorithm, has been chosen here to solve the inverse elastic problem with non-negativity constraints on the tractions, specifically the implementation described in \cite{Cantarella2004}.

\begin{remark}
It is noteworthy that any algorithms implementing non-negative least squares solutions must be performed \emph{on-line}, i.e., when the displacement vector $D$ in the inverse elastic problem is known.

Although the BLOCK3 algorithm is proved to converge in a finite number of iterations, the complexity of the involved operations (and therefore, the problem of putting real-time constraints on them) may prove a limitation in the case where hard real-time constraints are required.
\end{remark}

\section{Experimental Validation}
\label{sec:exp_validation}

\subsection{Implementation Details}
\label{sec:implementation}

The approach presented in this paper has been developed as an open source software.
The current implementation is based on C++ and is built on top of the Skinware framework \cite{Youssefietal2011, Youssefietal2015a, Youssefietal2015b}.
In particular, we highlight the following features:
\begin{itemize}
\item both Boussinesq-Cerruti's and Love's formulations have been implemented;
\item the algorithms for tractions reconstruction and surface deflections have been designed considering a clear division of the offline and online parts;
\item the implemented algorithms allow for the reconstruction of forces using displacements, pressures using displacements, displacements using forces, and displacements using pressures;
\item for the inverse problem, variants of the algorithms exist, which enforce the non-negativity constraints on tractions. 
\end{itemize}

As it can be observed from the derivation of the influence coefficients (both for Boussinesq-Cerruti's and Love's approaches) above, the elements of $C$ depend only on the geometry of the problem.
This fact can be exploited to divide the solution of the reconstruction problem into an offline and online part, which is fundamental for practical reasons from a computational standpoint.
When the straight least-squares solution via the pseudo-inverse is used, both the original $C$ matrix and its pseudo-inverse $C^{+}$ can be computed before tactile data processing starts.
Otherwise, only the $C$ matrix can be computed in advance.

\begin{figure}[t]
\centering
\includegraphics[width=\columnwidth]{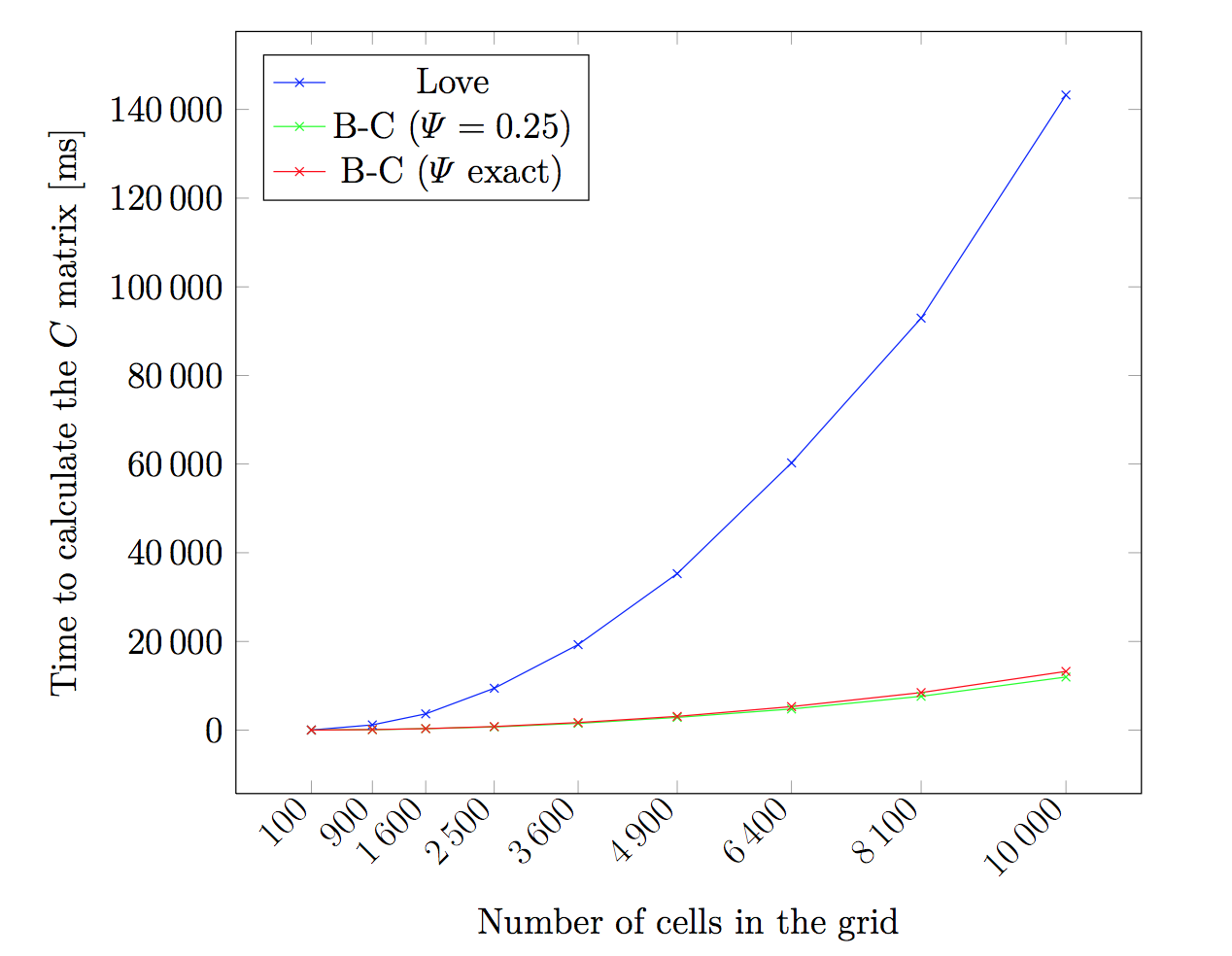}
\caption{Time taken to compute the influence coefficients matrix for different elastic models. Note the quadratic complexity with increasing grid size.}
\label{fig:time}
\end{figure}
This observation can be used to greatly improve the overall performance, since the only step to perform online would become a matrix-vector multiplication, which is a relatively lightweight operation.
In order to compare the time complexity of both the Boussinesq-Cerruti's and Love's solutions and to assess the gain obtained by offline computation, the $C$ matrix has been computed for different grid sizes.
Figure \ref{fig:time} shows computing times for an increasing number of grid cells.
Influence coefficients have been computed for two identical instances of regular square grids.
Only normal displacements and tractions have been considered.
Therefore, the size of the $C$ matrix is $n \times n$, where $n$ is the number of grid cells.

As expected, Figure \ref{fig:time} shows a quadratic time complexity for increasing grid sizes.
However, there is a large discrepancy in the growth rate for the Boussinesq-Cerruti's and Love's models.
Qualitatively, the time required to compute influence coefficients by the Love's model is tenfold the time for the Boussinesq-Cerruti's model in an equivalent grid.
This is possibly due to the high complexity of the formulas for computing Love's influence coefficients \eqref{eq:Lj} compared to the Boussinesq-Cerruti's solution \eqref{eq:uz}.
With such an increase in the computation time, the capability of performing the majority of calculations offline becomes crucial.

Two remarks should be made.
On the one hand, it is noteworthy that in case non-negativeness constraints for the reconstruction of tractions are considered, only the $C$ matrix can be pre-computed, whereas its pseudo-inverse cannot.
On the other hand, this division allows a number of computations to be performed \emph{off-site}, which is a great benefit if the tractions reconstruction were to be performed in real-time.
The software architecture could then enable solving the elastic problem with an adapting resampling resolution, querying external, off-site computation nodes for the required matrices.
This means that resource intensive computations could be moved to a dedicated unit, possibly endowed with dedicated hardware for maximum efficiency, e.g., a GPU, whereas the use of such hardware would be otherwise prohibitive on board the robot, e.g., for reasons of power efficiency or space occupation.

\begin{figure}[t]
\centering
\subfigure{
\includegraphics[width=1.22in]{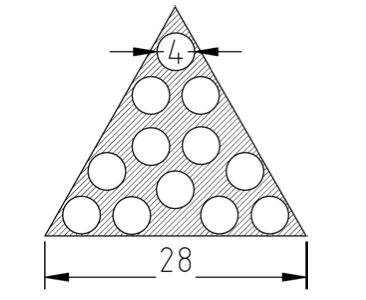}
\label{triangolo}}
\hspace{0.25mm}
\subfigure{
\includegraphics[width=1.2in]{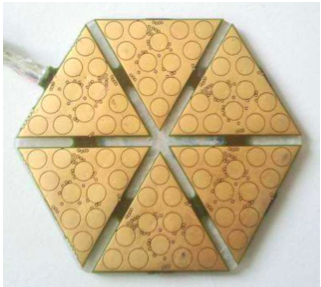}
\label{esagono}}
\caption{Six triangular modules (left) build up an hexagonal patch (right).}
\label{fig:step}
\end{figure}
\subsection{Experimental Setup}
\label{sec:experimental_setup}
Experimental data used to validate the presented models are taken from the dataset used in \cite{Muscari2013}. 
Data have been recorded using six triangular ROBOSKIN modules combined together to form a hexagonal patch, shown in Figure \ref{fig:step} on the right hand side\footnote{A video showing online contact shape reconstruction is available at \url{https://tinyurl.com/y7jc5goo}.}.
A three-way Cartesian robot positioner from Thorlabs Inc. allows the patch to horizontally translate along the $x$ and $y$ axes, and to rotate it along the vertical axis.
A force of up to $3$ $N$ can be applied using a linear actuator mounted along the vertical axis.
The tip of the linear actuator, which is provided with a load cell to measure the peak force on the area below, can be mechanically coupled with indenters of various shapes. 

\begin{figure}[t]
\centering
\includegraphics[width =\columnwidth]{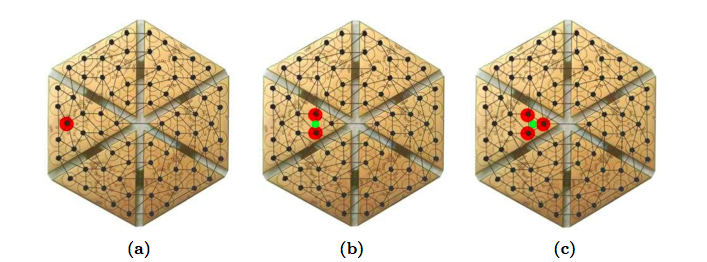}
\caption{Load conditions in different trials: indenter positioned over one taxel (left), between two taxels (centre) and between three taxels (right).}
\label{posizioni}
\end{figure}
Four parameters have been varied during the experimental campaign, namely the location of the pressure exerted by the rigid indenter (i.e., over a given taxel, Figure \ref{posizioni} on the left, between two taxels, Figure \ref{posizioni} in the centre, and between three taxels, Figure \ref{posizioni} on the right), the indenter's shape (i.e., a $12$ $mm$ diameter half-sphere, a $6$ $mm$ diameter half-sphere, a $12$ $mm$ diameter cylinder, a $3$ $mm$ diameter cylinder), the value of the exerted pressure (i.e., $0.2$ $N$, $0.5$ $N$, $1$ $N$, $1.8$ $N$, $2.5$ $N$, and $3$ $N$), and the duration of the contact phase (from $3$ $sec$ to $7$ $sec$).
Each trial involving any combinations of these parameters has been repeated $50$ times for statistical significance.

\begin{figure}[t]
\centering
\subfigure[Raw tactile data]{
\includegraphics[width=1.62in]{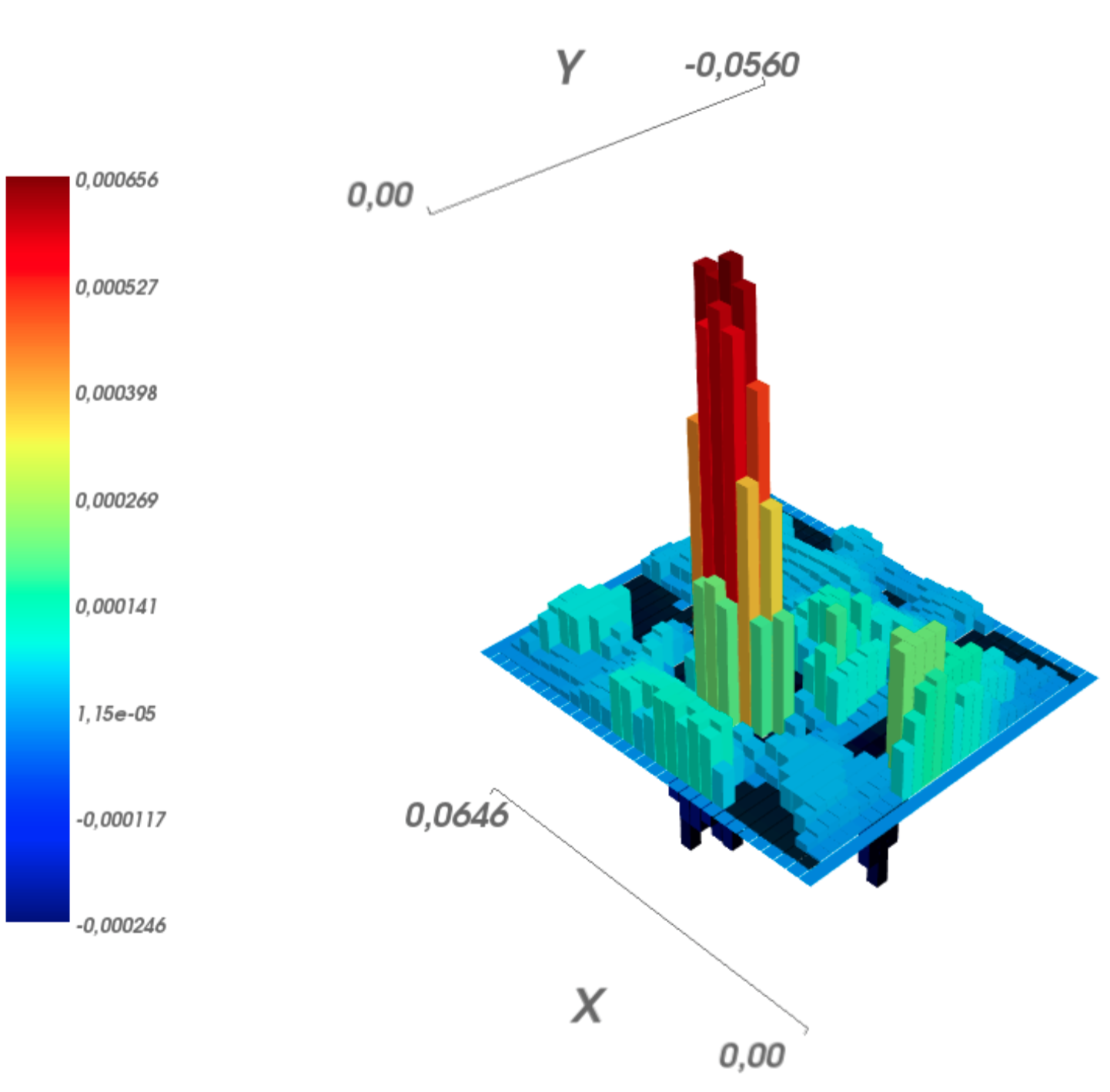}
\label{fig:bcraw}}
\subfigure[$0.5$ $mm$ cell size]{
\includegraphics[width=1.62in]{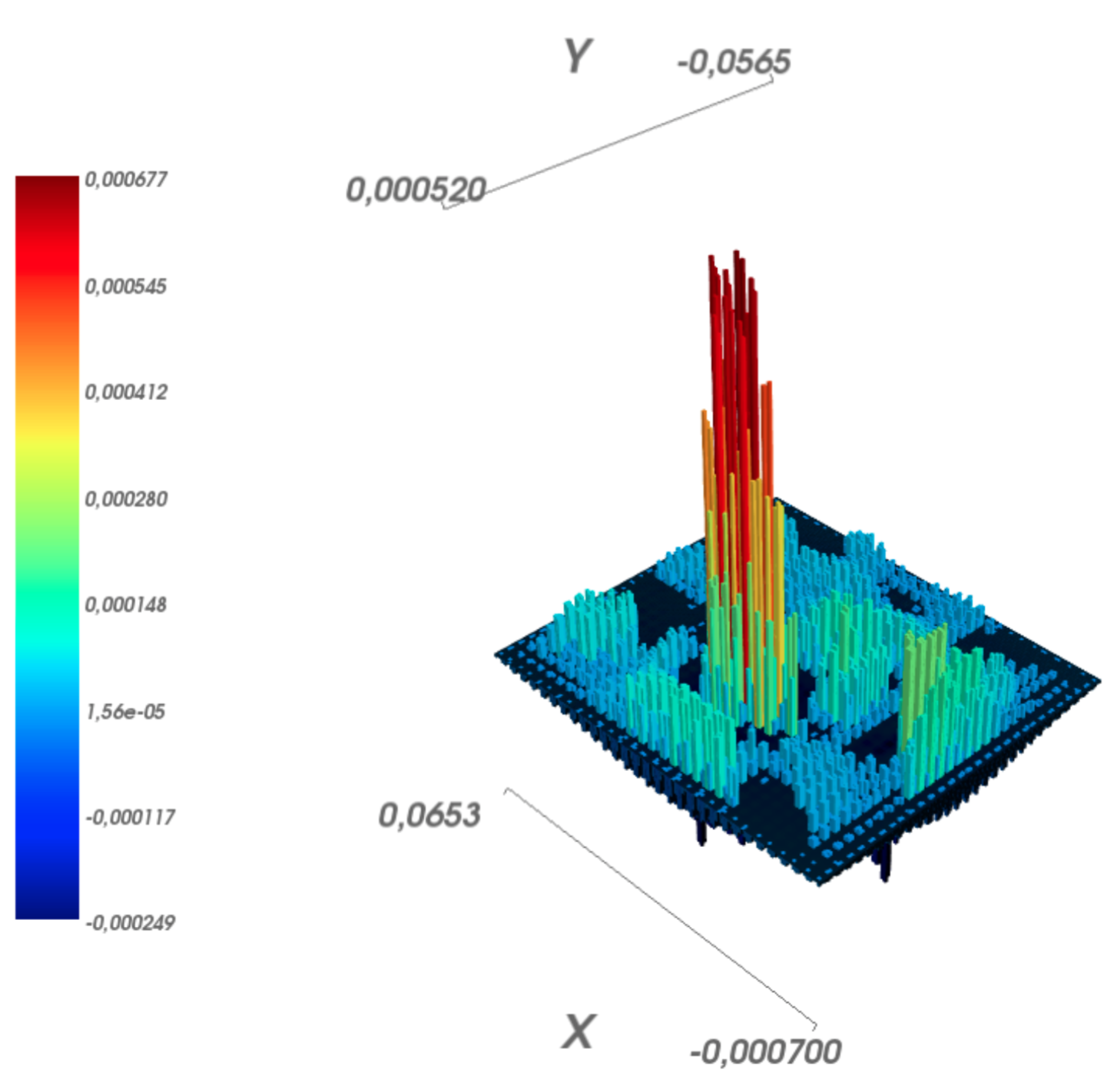}
\label{fig:bc05}}
\\
\subfigure[$2$ $mm$ cell size]{
\includegraphics[width=1.62in]{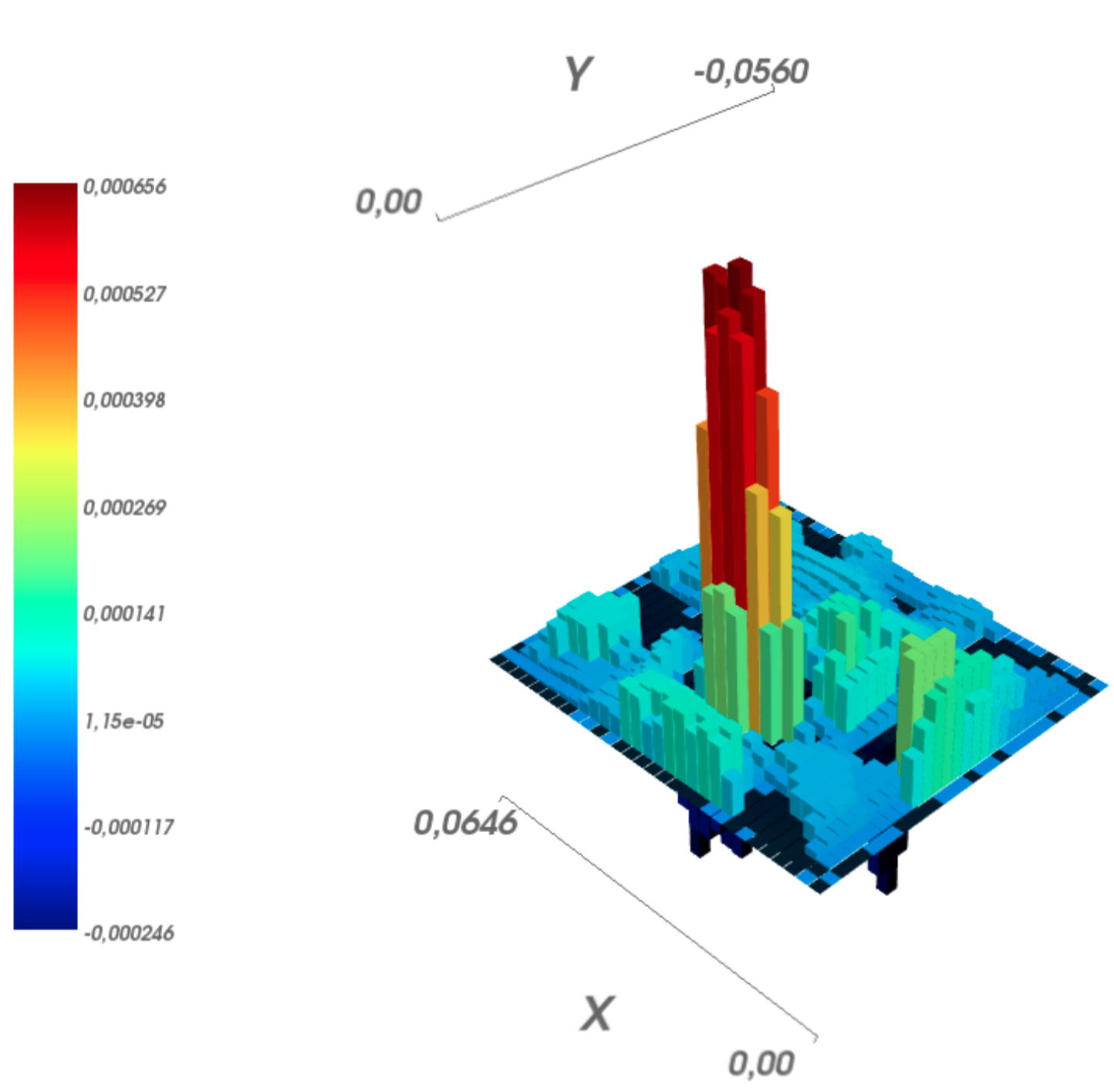}
\label{fig:bc2}}
\subfigure[$3$ $mm$ cell size]{
\includegraphics[width=1.62in]{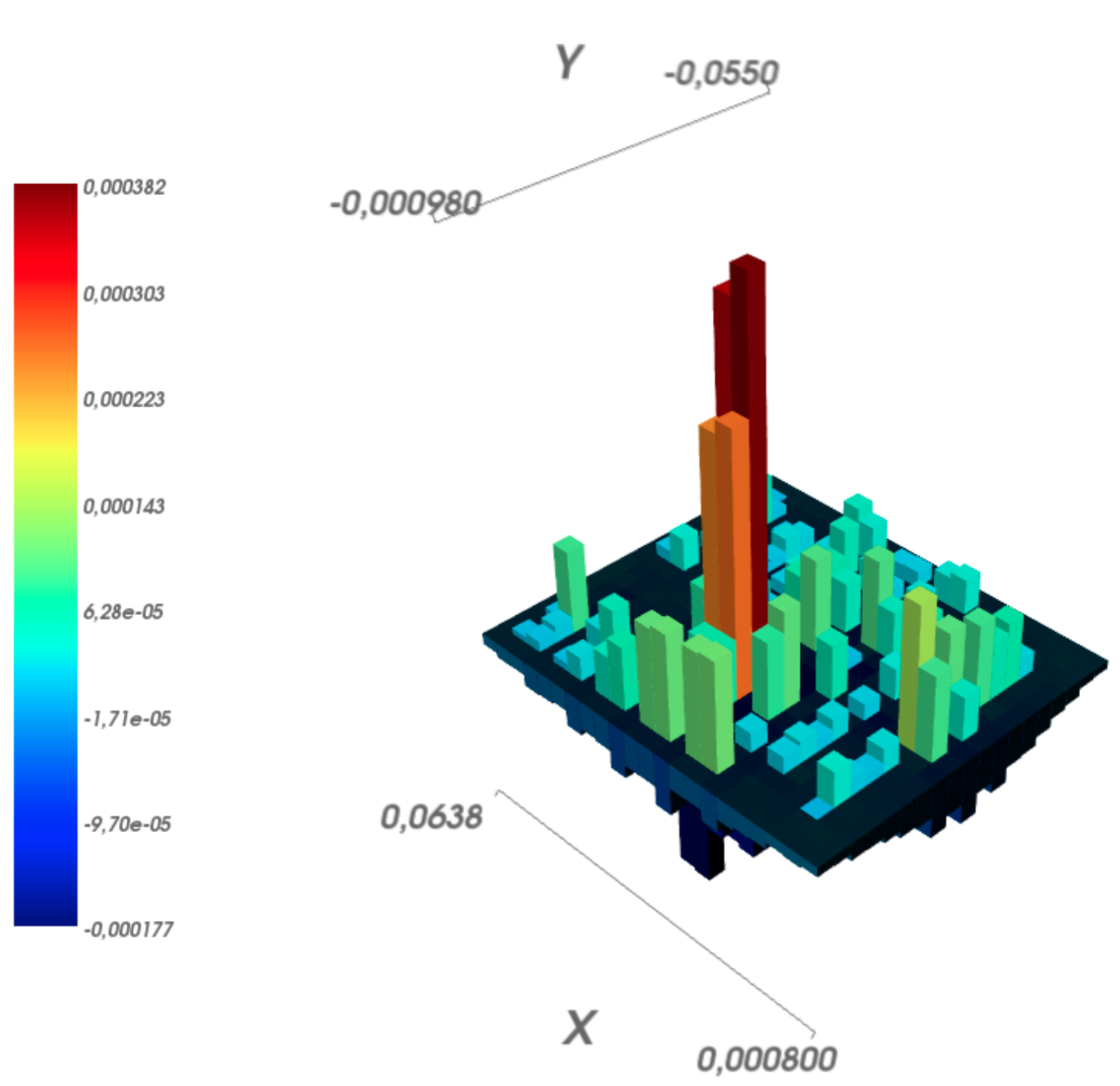}
\label{fig:bc3}}
\caption{Reconstruction of surface deflections according to Boussinesq-Cerruti's solution with a varying resolution of the reconstruction grid.}
\label{fig:deflectionBC}
\end{figure}

\begin{figure}[t]
\centering
\subfigure[B-C tractions]{
\includegraphics[width=1.62in]{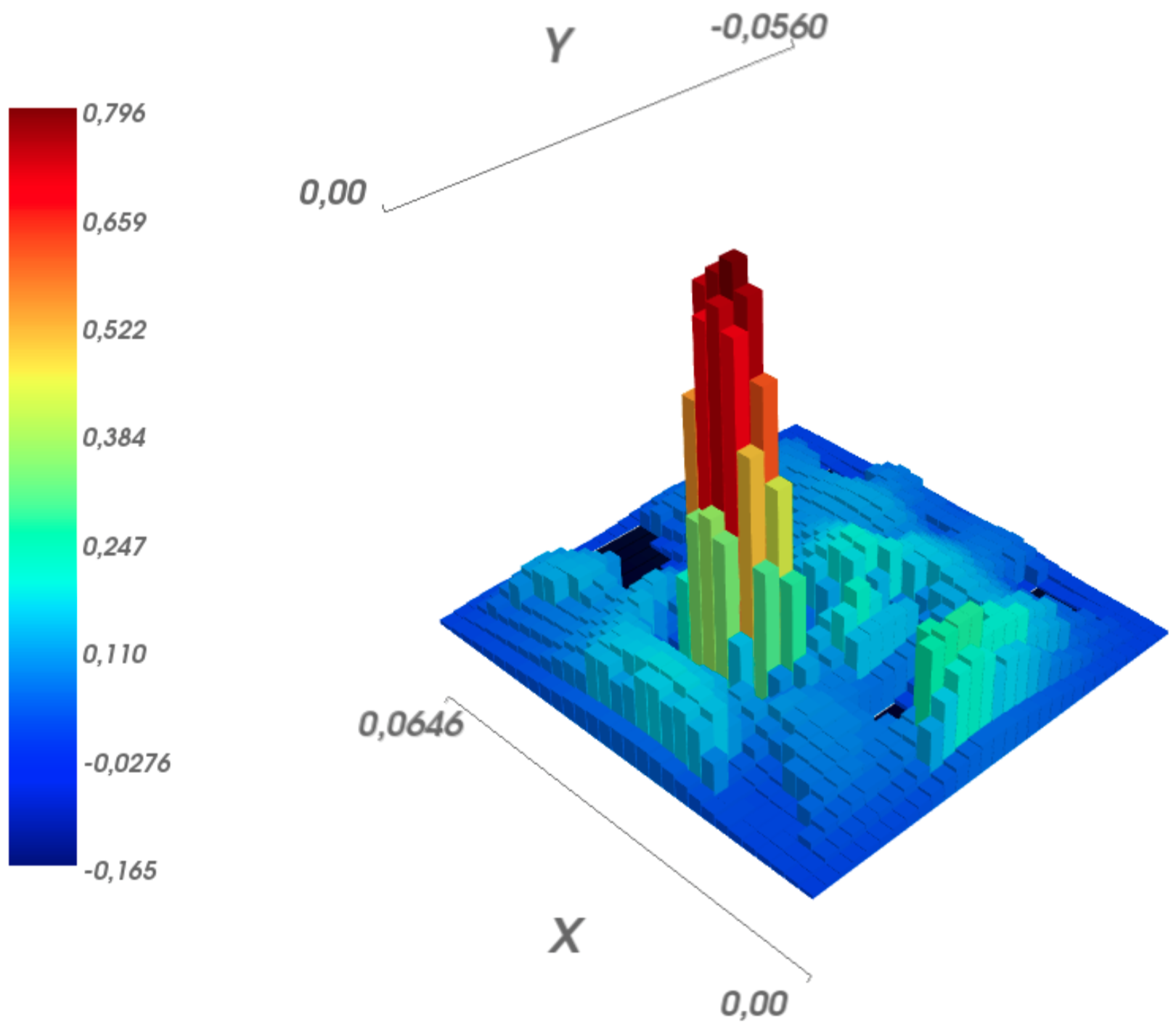}
\label{fig:bctraction}}
\subfigure[Love tractions]{
\includegraphics[width=1.57in]{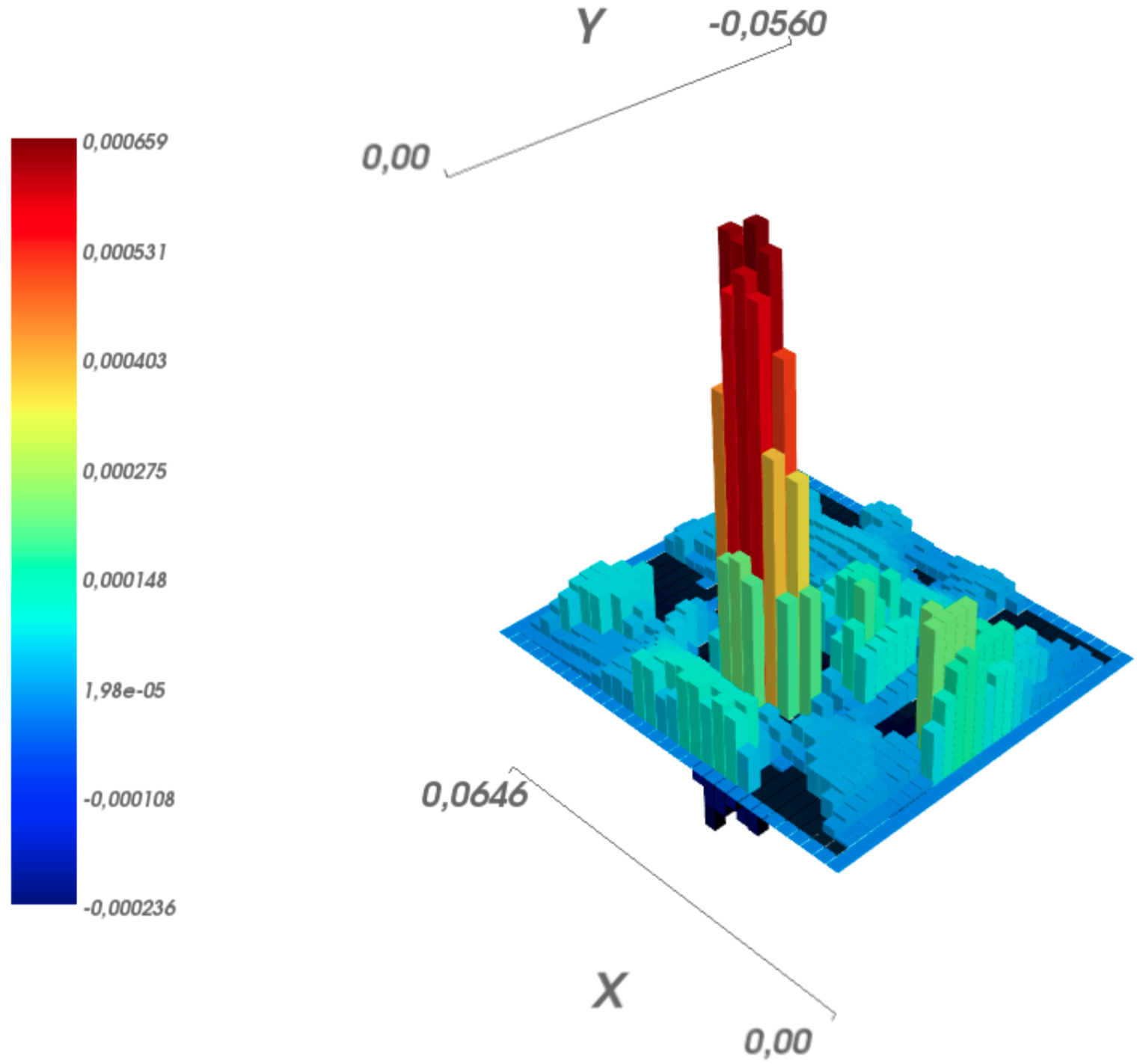}
\label{fig:lovetraction}}
\caption{Solutions to a \emph{square} inverse elastic problem, as given by Boussinesq-Cerruti's and Love's models.}
\label{fig:traction}
\end{figure}

\begin{figure}[t]
\centering
\subfigure[Raw tactile data]{
\includegraphics[width=1.62in]{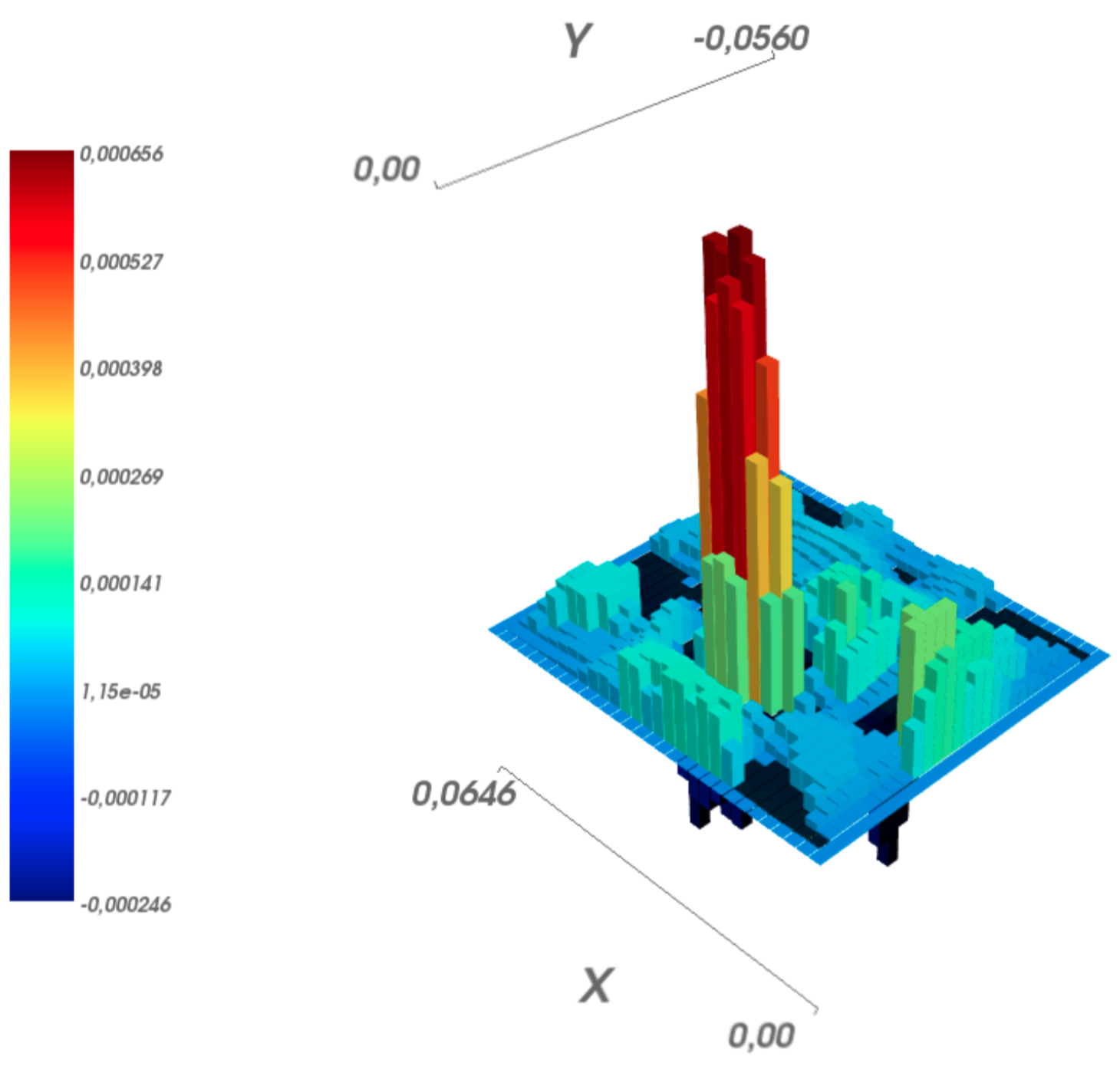}
\label{fig:loveraw}}
\subfigure[$0.5$ $mm$ cell size]{
\includegraphics[width=1.62in]{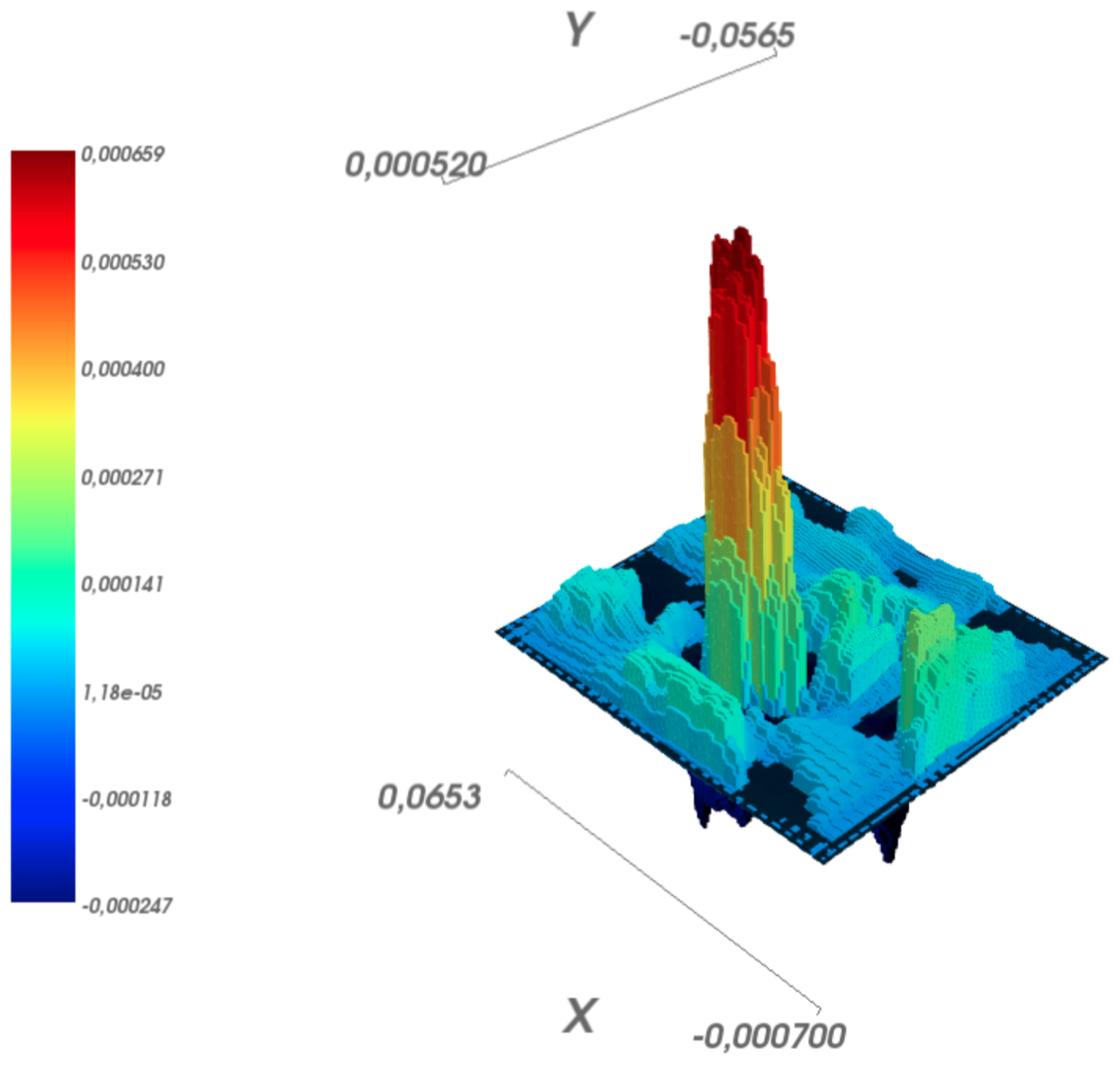}
\label{fig:love05}}
\\
\subfigure[$2$ $mm$ cell size]{
\includegraphics[width=1.62in]{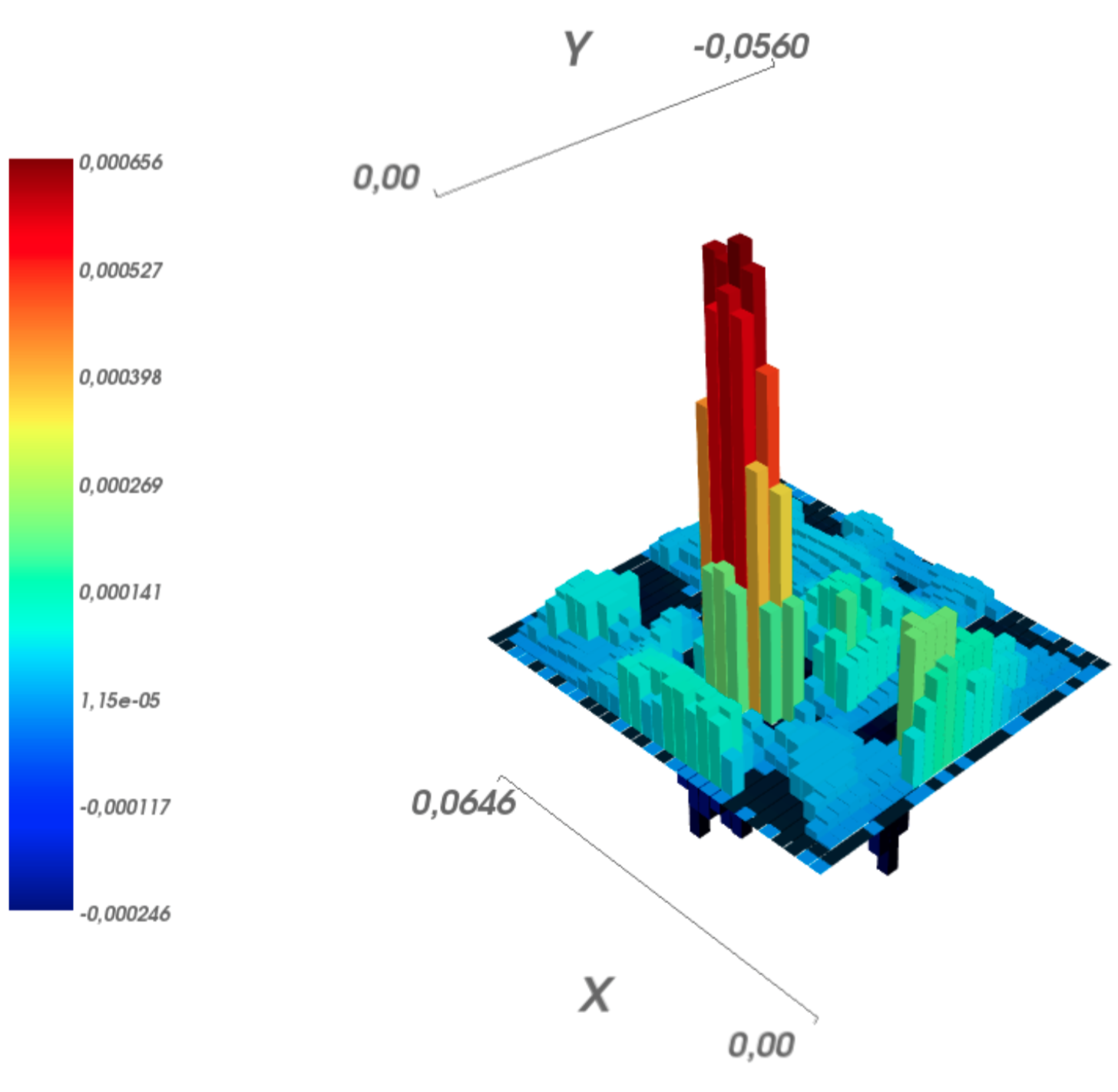}
\label{fig:love2}}
\subfigure[$3$ $mm$ cell size]{
\includegraphics[width=1.62in]{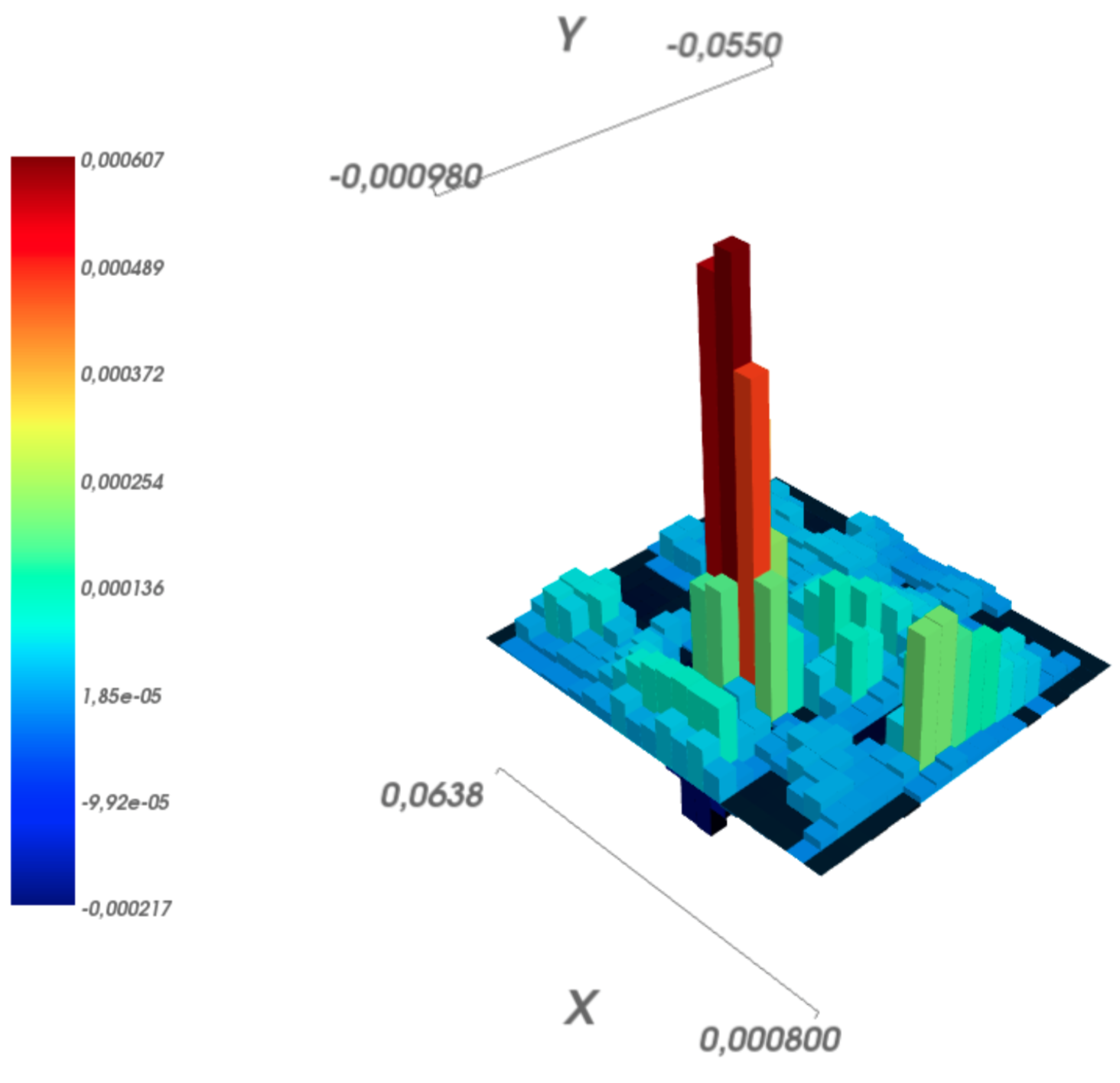}
\label{fig:love3}}
\caption{Reconstruction of surface deflections according to Love's solution with a varying resolution of the reconstruction grid.}
\label{fig:deflectionLove}
\end{figure}

\subsection{Performance of Love's Solution Compared to Boussinesq-Cerruti's Solution}
\label{sec:performance_comparison}

In all the experiments, Boussinesq-Cerruti's models have been computed using an approximation for the function $\Psi$ by a constant number, i.e., $\Psi = 0.25$, as suggested in \cite{Muscari2013}.
As a reference for further considerations, let us consider the experiment in which the robot skin patch is indented using a large sphere with a force of $1.8$ $N$, which corresponds to Figure \ref{fig:bcraw}.
The resolution of the grid over which the deflections are measured is about $2$ $mm$, see Figure \ref{fig:bc2}.

A qualitative comparison of results obtained for the inverse elastic problem using the Boussinesq-Cerruti's and Love's models is shown in Figure \ref{fig:traction}. 
The reconstruction grid is squared, i.e., the number of tractions to reconstruct is equal to the number of available measurement data points.
The two results are surprisingly similar, if we ignore the scale, and only minor differences can be noticed.
This is most probably due to the fine discretisation of the grid.
Each pressure acts on a $2 \times 2$ $mm$ size cell, which is in oversampling with respect to the actual ROBOSKIN spatial density (i.e., $2$ taxels per $cm^2$).

To compare the reconstruction quality for the two solutions, surface deflections have been reconstructed:
\begin{enumerate}
\item determining the tractions by solving a square inverse elastic problem in which the grids have the exact same geometry and are placed one over another;
\item reconstructing surface deflections for varying resolutions of the reconstructed displacements grid, namely (i) a grid with cells shaped as squares with a side length equal to $0.5$ $mm$, (ii) a grid being an exact copy of the input sensor data grid, and (iii) a grid with square shaped cells with side length equal to $3$ $mm$.
\end{enumerate}
Results are shown in Figure \ref{fig:deflectionBC} for the Boussinesq-Cerruti's solution and in Figure \ref{fig:deflectionLove} for the Love's solution.

Several comments can be made.
The first comment is that both in the Boussinesq-Cerruti's and Love's cases, reconstructing surface deflections with respect to a grid that is an exact copy of the sensory input grid, results in surface deflections which are exactly equal to the sensor input.
Obviously enough, this is to be expected.
In both cases, the influence coefficient matrix for the inverse problem $C_{I}$ is square.
Moreover, the coefficient matrix describing the forward problem $C_{F}$ is square as well, and most importantly those two matrices are equal. 
Therefore, in this particular case, reconstructing effective surface displacements $D$ corresponds to:
\begin{equation}
D = C_{F} \cdot Q = C_{F} C^{-1}_{I} \cdot S = I \cdot S 
\label{eq:deflection}
\end{equation}
This effect is evident when one compares Figures \ref{fig:bcraw} and \ref{fig:bc2}, and Figures \ref{fig:loveraw} and \ref{fig:love2}, for Boussinesq-Cerruti's and Love's solutions, respectively.

Then, it must be remarked that Boussinesq-Cerruti's solutions perform poorly in reconstructing surface displacements for grids characterised by a resolution different from the input's, as shown in Figures \ref{fig:bc05} and \ref{fig:bc3}.
On the one hand, it is apparent that Boussinesq-Cerruti's solutions perform well for problems in which the influence coefficient matrix is square, but once this property is lost, undesired artefacts appear.
Therefore, we conclude that Boussinesq-Cerruti's solutions are not suitable for a stable resampling in the displacements field.
On the other hand, if we consider the performance and stability of Love's solutions in reconstructing surface displacements for grids with a different resolution than the input, i.e., resampling the displacements field, shown in Figures \ref{fig:love05} and \ref{fig:love3}, one immediately notes the consistency of the reconstruction.
For an increased resolution of the reconstruction, as shown in Figure \ref{fig:love05}, Love's solutions yield a \emph{smooth} interpolation of sensor data, which is close to what one would expect to happen in actual robot skin.
If the Love's solution is used for downsampling the surface deflection field, as shown in Figure \ref{fig:love3}, the result is a deflection field which is still representative of the contact event.
It is noteworthy that the difference between the height of the main peak in the input and those of down-sampled cases is of the order of one twentieth of a millimetre.

\begin{figure}[t]
\centering
\subfigure[Tractions at $3$ $mm$ with B-C]{
\includegraphics[width=1.64in]{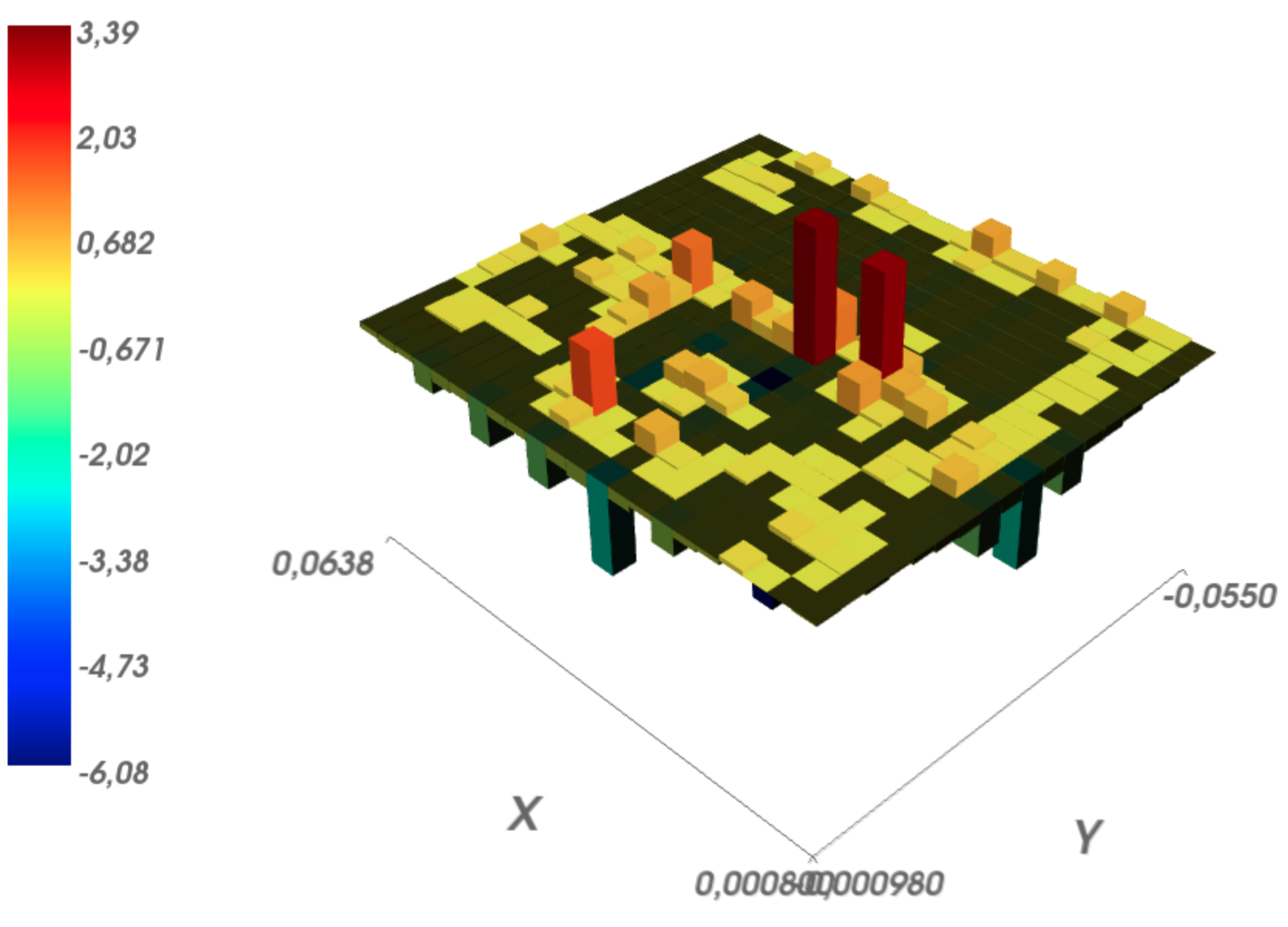}
\label{fig:3mmBC}}
\subfigure[Tractions at $3$ $mm$ with Love]{
\includegraphics[width=1.59in]{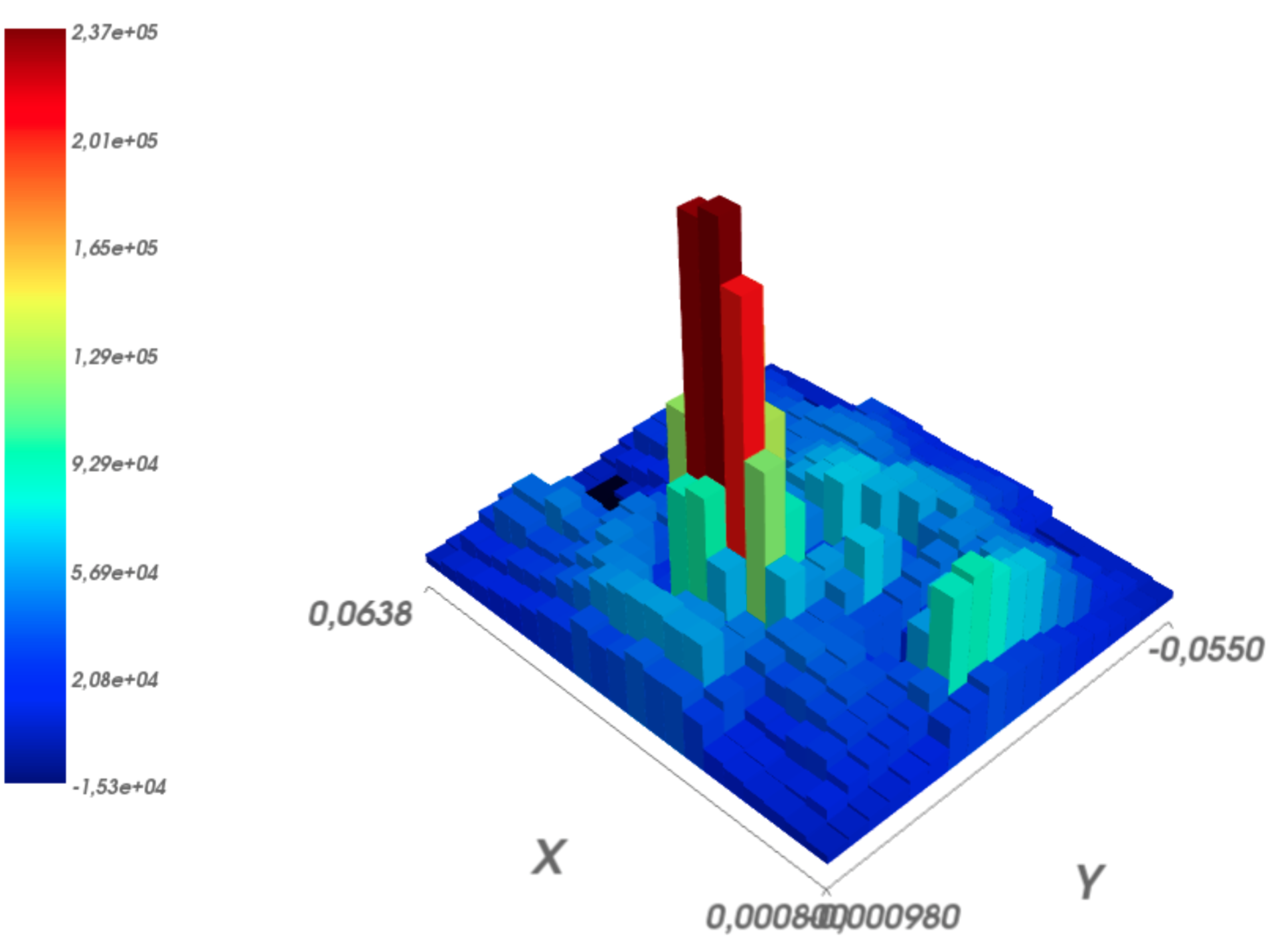}
\label{fig:3mmlove}}
\\
\subfigure[Reconstr. at $2$ $mm$ with B-C]{
\includegraphics[width=1.63in]{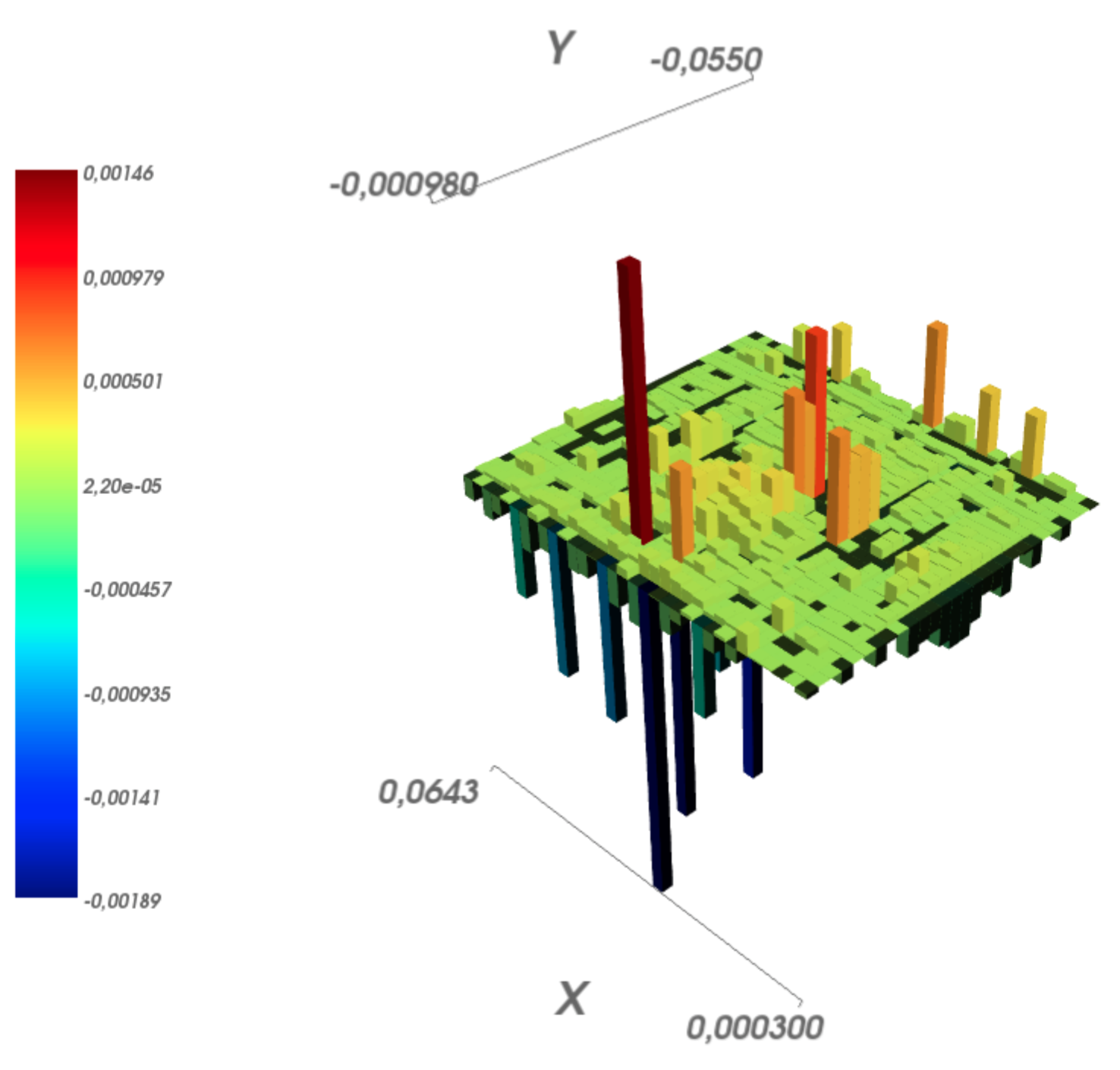}
\label{fig:2mmBC}}
\subfigure[Reconstr. at $2$ $mm$ with Love]{
\includegraphics[width=1.60in]{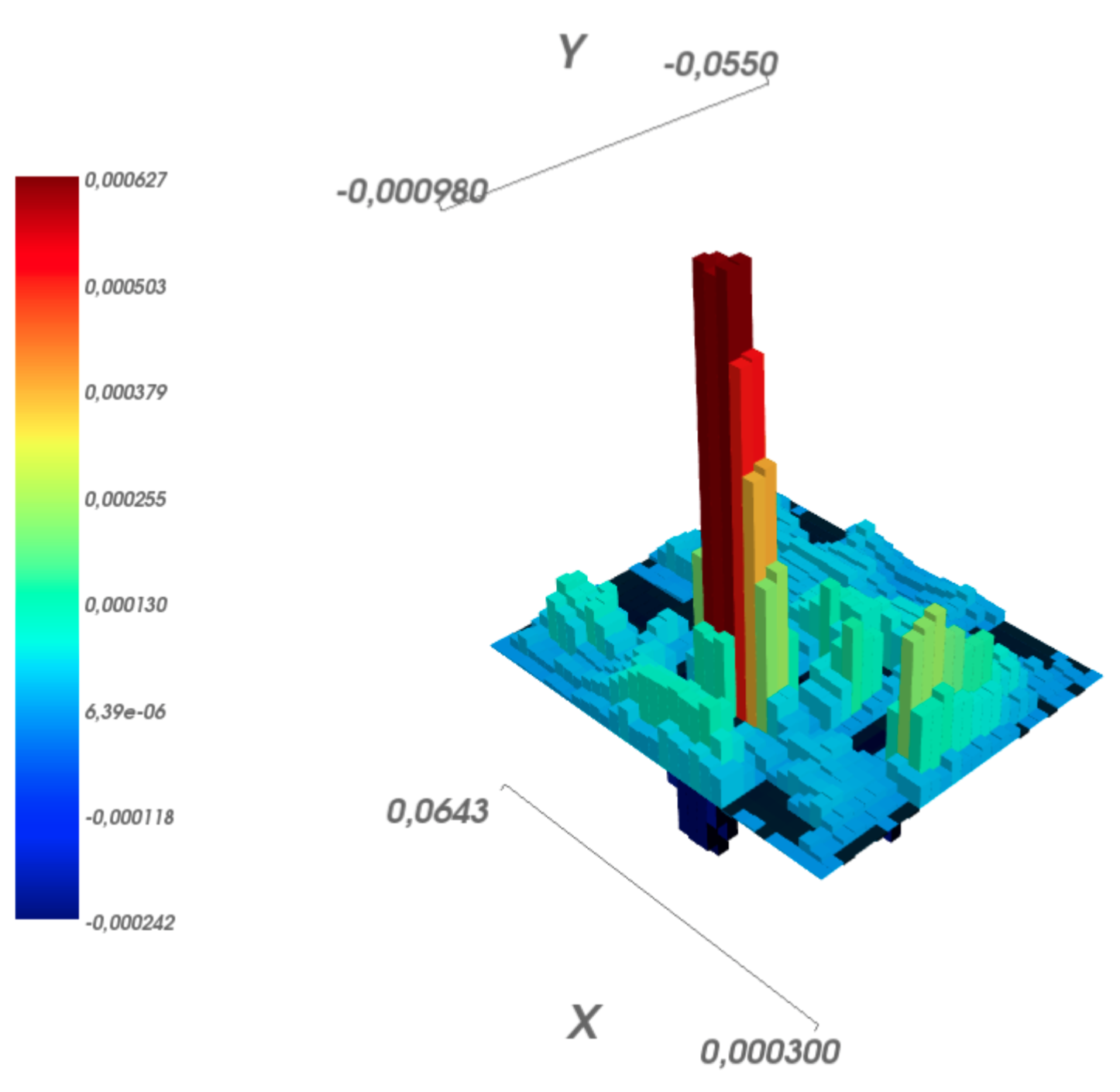}
\label{fig:2mmlove}}
\caption{Downsampling of the tractions grid and its influence on the reconstruction. In both cases, tractions are reconstructed on a regular square grid with $3$ $mm$ cell resolution. The forward elastic problem is computed for displacements reconstruction on a regular square $2$ $mm$ size grid.}
\label{fig:downsampling}
\end{figure}

%
The difference between the two models is even more apparent if a downsampling of the tractions field is considered, as shown in Figure \ref{fig:downsampling}.
In the Boussinesq-Cerruti's case, if the tractions are reconstructed on a regular grid with $3$ $mm$ grid size square cells, as shown in Figure \ref{fig:3mmBC}, the result is hard to interpret at best.
In addition, the reconstructed surface deflections, as shown in Figure \ref{fig:2mmBC}, do not resemble in any way the raw input sensor data shown in Figure \ref{fig:bcraw}.
On the contrary, Love's solution performs better in both tractions reconstruction and surface deflections, as exemplified in Figures \ref{fig:3mmlove} and \ref{fig:2mmlove}.
In particular, the similarity between Figures \ref{fig:2mmlove} and \ref{fig:bc3} must be noted.
The latter Figure corresponds to reconstructing the displacements field at $3$ $mm$, without downsampling.
It shows that about the same amount of \emph{information} or \emph{precision} is lost regardless whether the tractions grid or the reconstructed surface deflections grid is down-sampled.
This argument is backed by the dissimilarity of Figures \ref{fig:bc2} and \ref{fig:2mmlove}, which carry a different amount of information regardless of the fact that the reconstructed displacement grids are (approximately) of the same resolution.

On the one hand, the above observations can be exploited when using the algorithms for an actual reconstruction to cut down the overall computational cost.
If the aim is to down-sample the reconstructed displacements field, instead of solving an exact inverse elastic problem and then a smaller forward elastic problem, one can solve both problems at a smaller scale, since about the same amount of information is preserved, and therefore saving precious computational time.
On the other hand, increasing the actual resolution of the reconstructed \textit{tactile image} by resampling the tractions field has not been explored in this work.
It is noteworthy that there exists an infinity of solutions and not well-informed principles have been identified to single one out.

\begin{figure}[t]
\centering
\subfigure[Tractions, $n \times n$ grid, B-C]{
\includegraphics[width=1.64in]{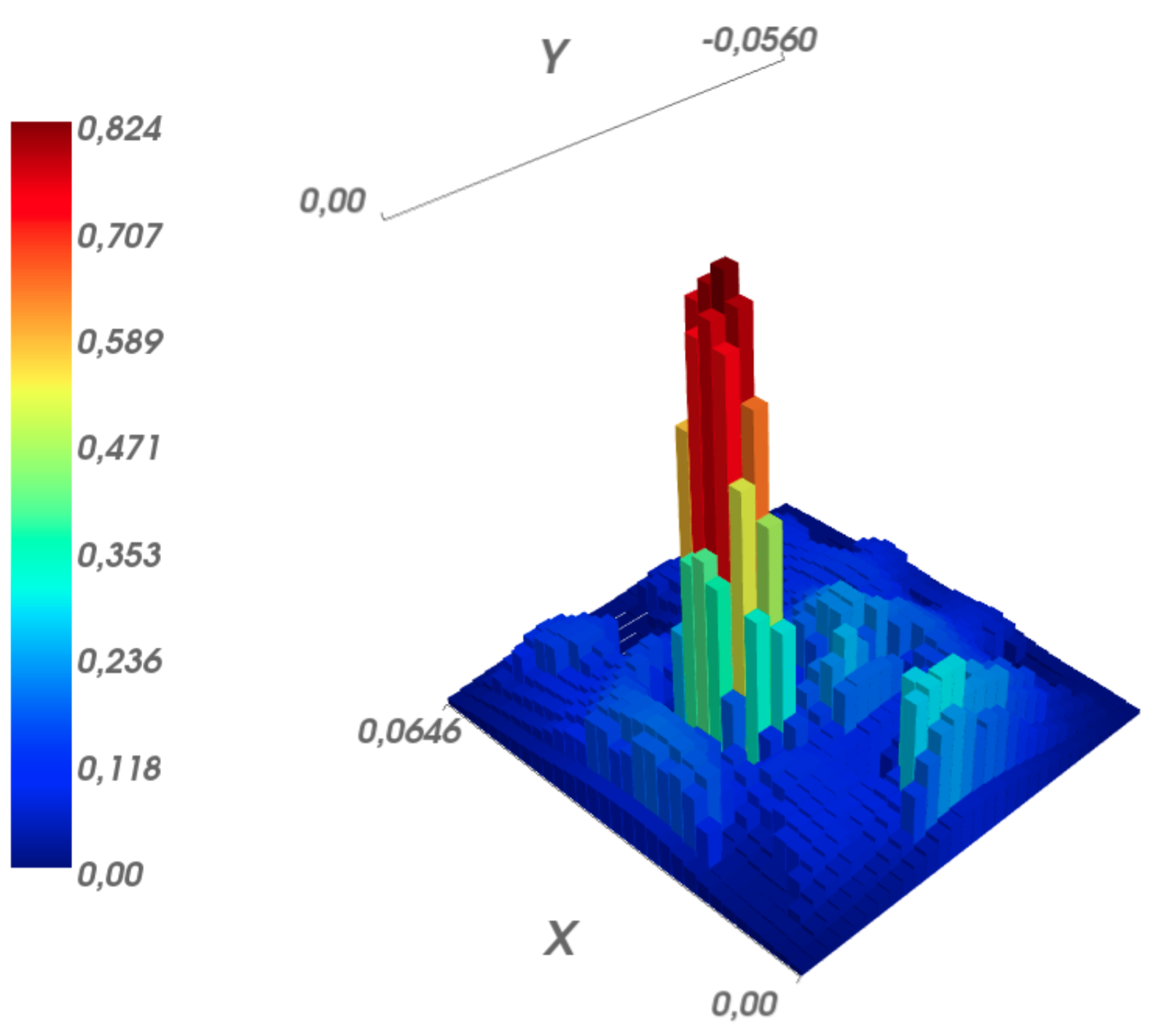}
\label{fig:nnbctraction}}
\subfigure[Tractions, $n \times n$ grid, Love]{
\includegraphics[width=1.61in]{lovetraction}
\label{fig:nnlove}}
\\
\subfigure[Deflections, $n \times n$ grid, B-C]{
\includegraphics[width=1.62in]{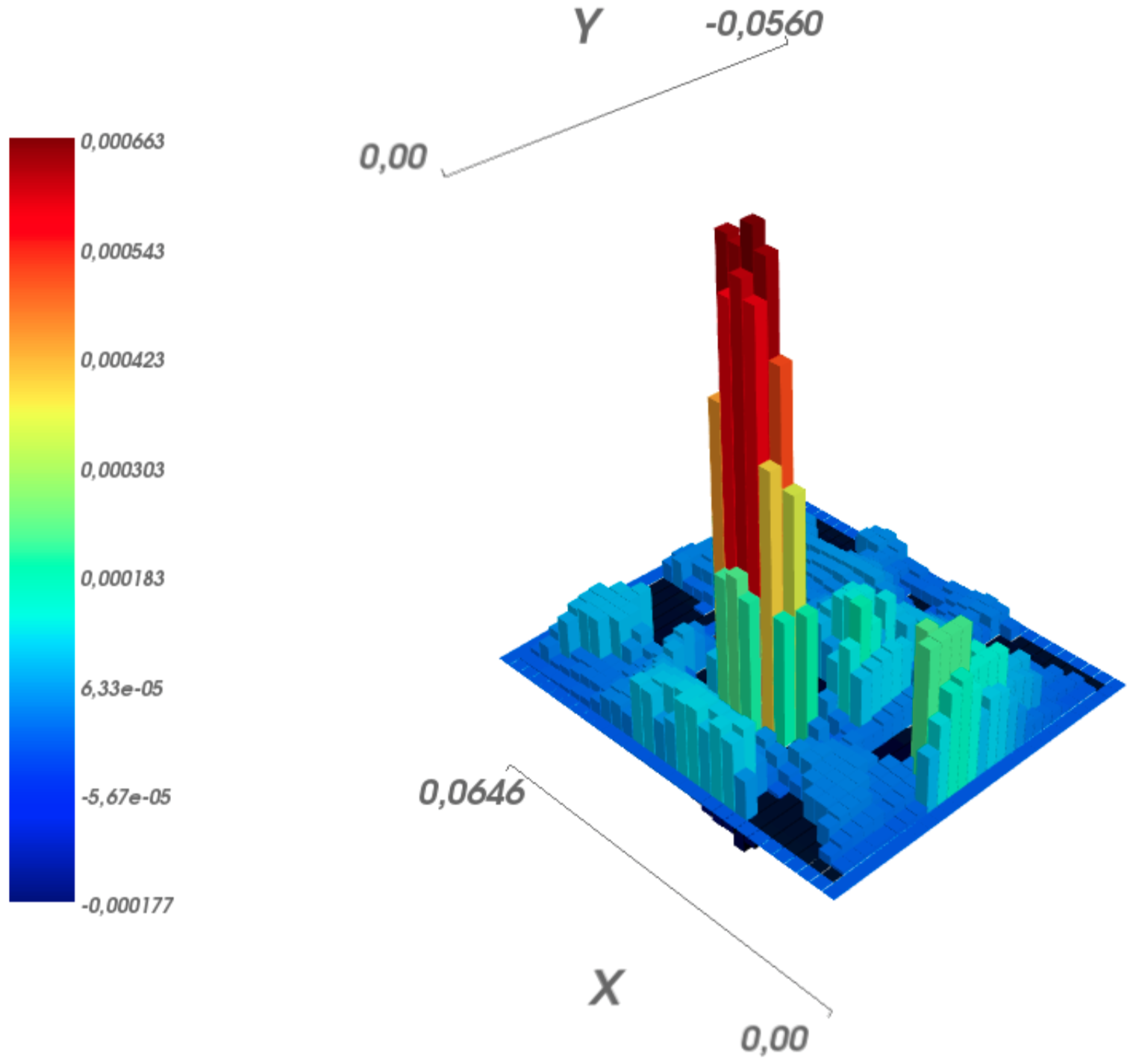}
\label{fig:nnbcdef}}
\subfigure[Deflections, $n \times n$ grid, Love]{
\includegraphics[width=1.62in]{lovetraction}
\label{fig:nnlovedef}}
\caption{Influence of non-negativity constraints on traction and deflection reconstructions for Boussinesq-Cerruti's and Love's models.}
\label{fig:nntraction}
\end{figure}

As shown in Section \ref{sec:compressive_NNLS}, while solving an inverse elastic problem, one may restrict the set of traction components to be non-negative, by using one of the non-negative least squares algorithms available in the literature.
Figure \ref{fig:nntraction} shows examples of such results.
Specifically, it is noteworthy that the solutions obtained by imposing the non-negativity constraint are not very dissimilar to \textit{free} solutions, for example those shown in Figures \ref{fig:bctraction} and \ref{fig:lovetraction}.
Furthermore, the reconstructed surface deflections resemble sensor measurements from Figure \ref{fig:bcraw} with a high degree of fidelity.
The already high fidelity degree of \textit{free} Love's solutions, as exemplified in Figure  \ref{fig:deflectionLove}, arises a question of usefulness of non-negativity constraints in the first place.
The constraint does improve the solution by eliminating pathological components.
However, it makes meeting hard real-time requirements harder, due to the complexity of the involved algorithms, which can be only run online.
Furthermore, the online phase, which solves a system of linear equations with non-negativity constraints, is expected to be slower than in the case of a \textit{free} solution, which involves merely a single matrix-vector multiplication operation.

\begin{figure}[t]
\centering
\subfigure[Tractions, B-C, exact $\Psi$]{
\raisebox{0.15\height}{
\includegraphics[width=1.63in]{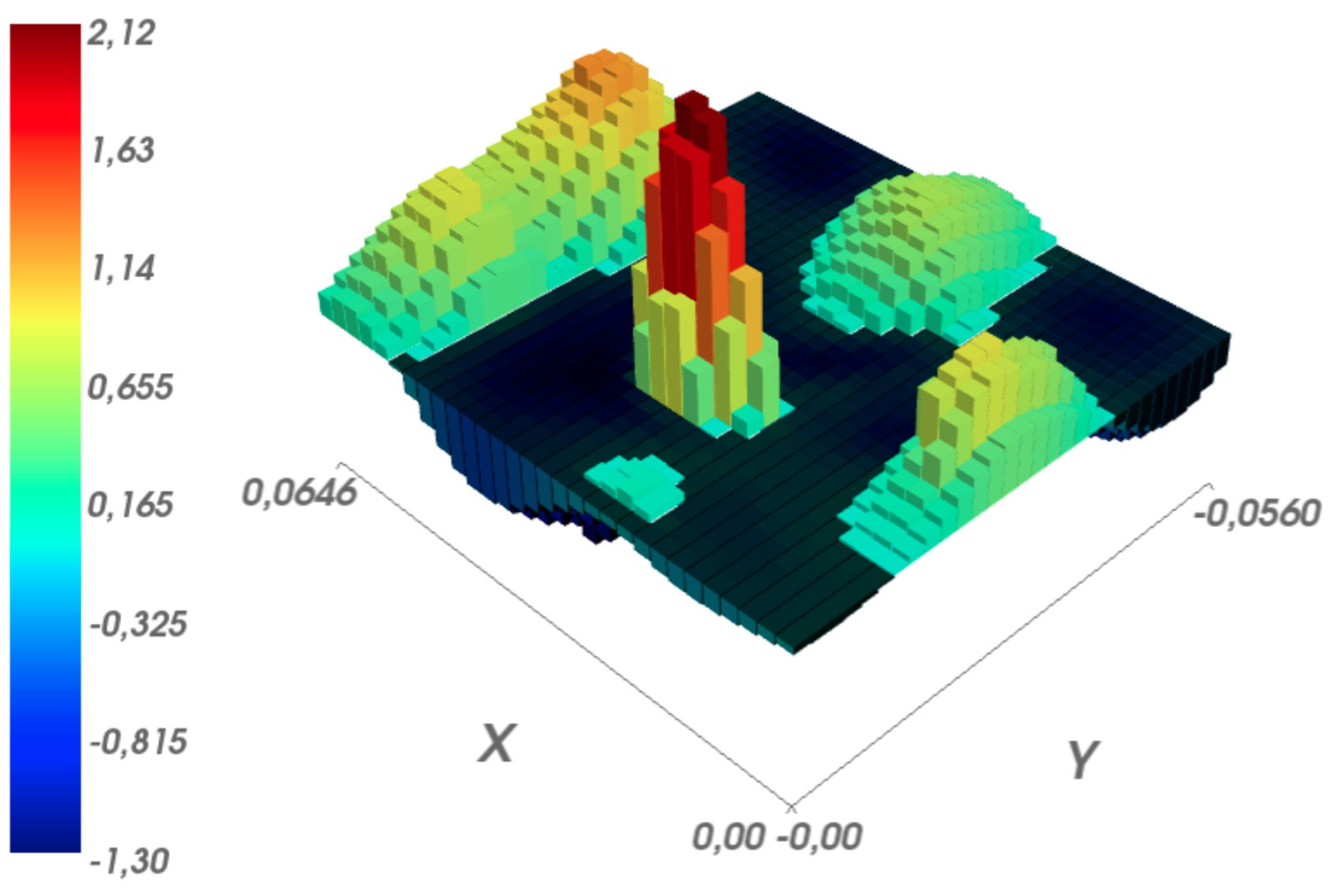}}
\label{fig:bctract12}}
\subfigure[Tractions, B-C, $n \times n$, exact $\Psi$]{
\includegraphics[width=1.57in]{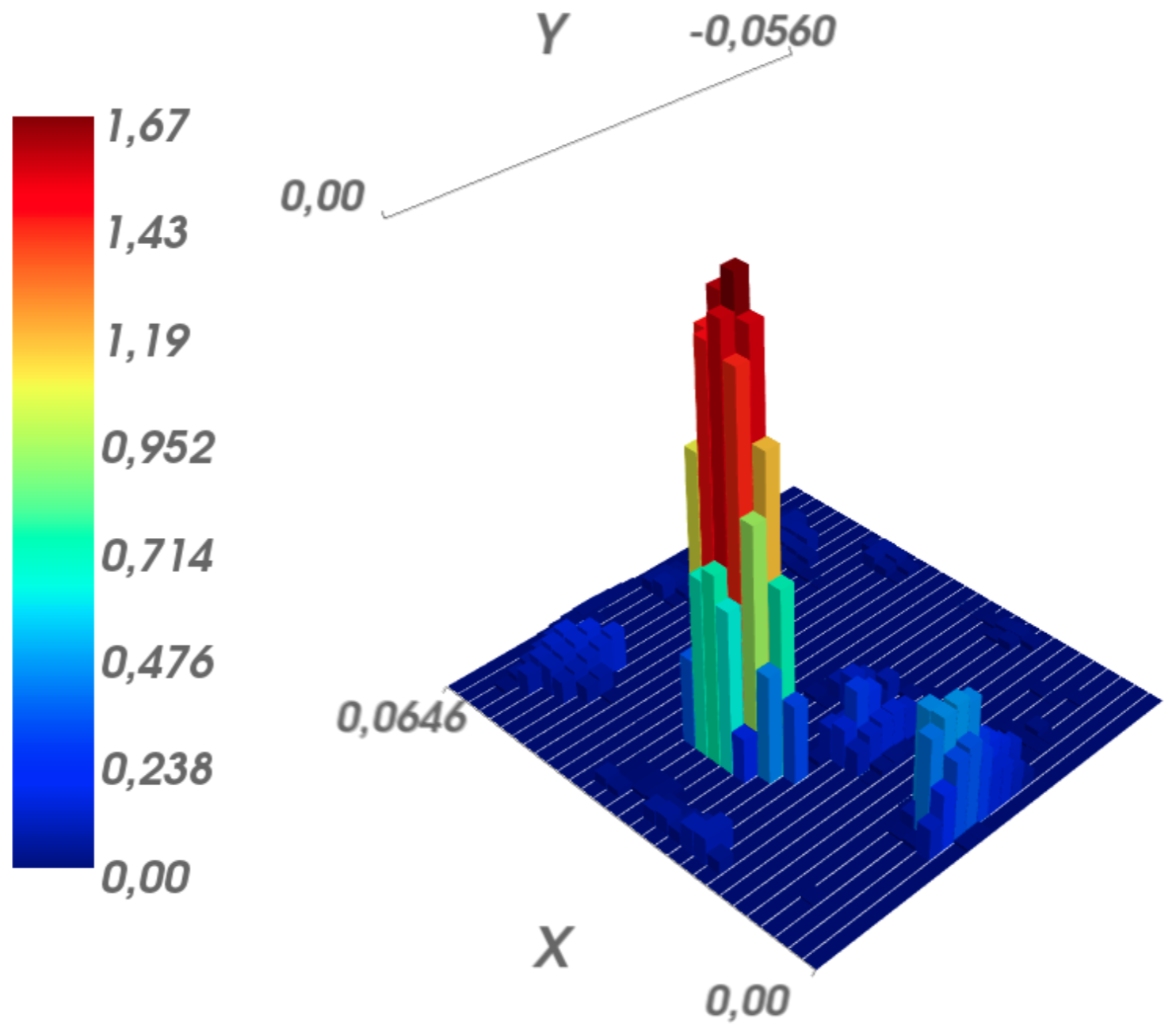}
\label{fig:nnbctract12}}
\\
\subfigure[Deflections, B-C, exact $\Psi$]{
\includegraphics[width=1.62in]{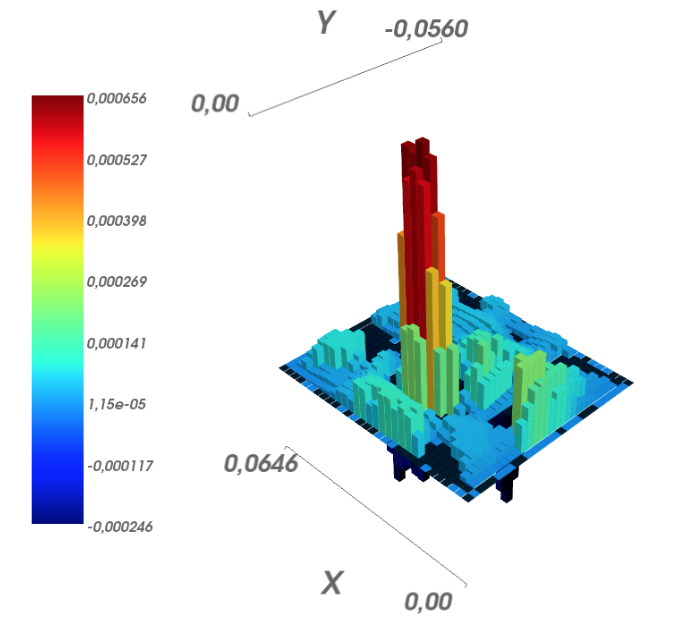}
\label{fig:bcdef12}}
\subfigure[Deflections, B-C, $n \times n$, exact $\Psi$]{
\includegraphics[width=1.62in]{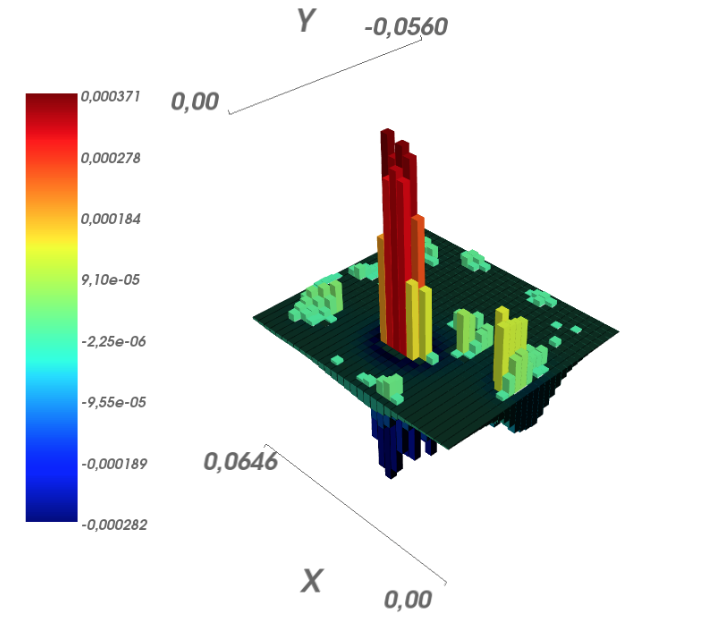}
\label{fig:nnbcdef12}}
\caption{Results in computing the Boussinesq-Cerruti's solution with the exact formulation for the $\Psi$ function.}
\label{fig:dodici}
\end{figure}
As shown in Section \ref{sec:Boussinesq_solution}, the $\Psi$ function in the approximate Boussinesq-Cerruti's solution can be approximated by a fixed number, i.e., $\Psi(x) \approx 0.25$.
The authors of \cite{Muscari2013} have explicitly set $\Psi$ to this fixed value.
To explore the impact of this choice, traction reconstruction with the exact calculation of the $\Psi$ function has been performed, using both the \textit{free} and the non-negativity-constrained solutions.
The result is shown in Figure \ref{fig:dodici}.
The free solution, shown in Figure \ref{fig:bctract12}, still shows a peak in the contact location, but nevertheless contains a very large amount of noise, the interpretation of which is difficult.
Obviously enough, for the reasons backed by \eqref{eq:deflection}, the reconstruction of surface displacements corresponds exactly to input data. 
The non-negativity-constrained solution presents an interesting filtering behaviour.
Forces are purported to be exhibited only in the location of highest surface deflections.
The surface deflection reconstruction is not characterised by the fidelity with respect to input data as the Love's solution.
However, it is possible that such reconstruction ($\Psi$ computed using the exact formula along with non-negativity constraints) effectively filters noise from the solution with respect to input data.
Possibly, the areas with the largest discrepancies between said reconstruction and input data are those where the strongest noise is prevalent.
Nevertheless, without a detailed analysis of noise sources and their nature, this statement cannot be defended trivially.

\section{Conclusions}
This paper introduces an approach to extract information about contact events based on large-scale tactile sensors.
The approach is based on a closed-form algorithm for the reconstruction of the contact shape, which is backed by a physical model of a large-scale, capacitance-based robot skin technology we have developed in the past few years, namely ROBOSKIN.
The classical solution that can be obtained by adopting the Boussinesq-Cerruti's model is compared (both from a qualitative and a computational perspective) to the Love's approach for solving a distributed inverse  contact problem, which has been adapted to our case.
The paper elaborates on two aspects: the first is a proposal for a general-purpose algorithm for the reconstruction of deformation and force distributions in case of capacitance-based robot skins; the second is related to the characterisation of its real-time performance, which can be tuned according to available computational resources.
Experiments on robot skin patches have been performed to provide a quantitative analysis of results.

\section*{Acknowledgements}
The research leading to these results has received funding from the European Community's Seventh Framework Programme (FP7/2007-2013) under Grant 231500 (project ROBOSKIN) and Grant 288553 (project CloPeMa).

\section*{References}
\bibliography{fulvio}

\end{document}